\documentclass[letterpaper, 10 pt, conference]{ieeeconf}  
\IEEEoverridecommandlockouts
\usepackage{amsmath,amssymb,xcolor}
\usepackage[]{graphicx}

\usepackage{algpseudocode}
\usepackage{algorithm}
\usepackage{hyperref}

\newcommand{\vv}{\mathbf{v}}

\newcommand{\vx}{\mathbf{x}}

\newcommand{\st}{\mbox{s.t.}}

\DeclareMathOperator*{\rank}{rank}

\DeclareMathOperator{\shrink}{shrink}

\DeclareMathOperator*{\argmin}{argmin}
\newcommand{\R}{\mathbb{R}}
\DeclareMathOperator*{\tr}{tr}
\DeclareMathOperator*{\diag}{diag}

\newcommand{\norm}[1]{\lVert#1\rVert}
\graphicspath{{./figs/}}

\title{\huge Robust Dual-Graph Regularized Moving Object Detection}

\author{\authorblockN{Jing Qin}
\authorblockA{\textit{Department of Mathematics}\\
\textit{University of Kentucky}\\
\textit{Lexington, KY 40506, USA}\\
\textit{jing.qin@uky.edu}\\}
\and
\authorblockN{Ruilong Shen, Ruihan Zhu and Biyun Xie}
\authorblockA{\textit{Department of Electrical and Computer Engineering}\\
\textit{University of Kentucky}\\
\textit{Lexington, KY 40506, USA}\\
\textit{\{Ruilong.Shen, Ruihan.Zhu, Biyun.Xie\}@uky.edu}\\}}

\begin{document}

\maketitle

\begin{abstract}
Moving object detection and its associated background-foreground separation have been widely used in a lot of applications, including computer vision, transportation and surveillance. Due to the presence of the static background, a video can be naturally decomposed into a low-rank background and a sparse foreground. Many regularization techniques, such as matrix nuclear norm, have been imposed on the background. In the meanwhile, sparsity or smoothness based regularizations, such as total variation and $\ell_1$, can be imposed on the foreground. Moreover, graph Laplacians are further imposed to capture the complicated geometry of background images. Recently, weighted regularization techniques including the weighted nuclear norm regularization have been proposed in the image processing community to promote adaptive sparsity while achieving efficient performance. In this paper, we propose a robust dual-graph regularized moving object detection model based on the weighted nuclear norm regularization, which is solved by the alternating direction method of multipliers (ADMM). Numerical experiments on body movement data sets have demonstrated the effectiveness of this method in separating moving objects from background, and the great potential in robotic applications.

\end{abstract}


\section{Introduction}

The development of advanced robotic technologies has released traditional robots isolated by fences or other protective barriers to environments with human beings \cite{zanchettin2019towards}. Such kinds of robots that are safe and intelligent enough to work alongside or directly interact with humans are called collaborative robots, including lightweight industrial robots, social robots, service robots, etc \cite{stein2020collaborative}. Human motion detection plays a significant role in the motion planning and control of collaborative robots to improve the safety and efficiency of human-robot interaction. On the one hand, the detected human motion will be used as the input information of various real-time motion planning algorithms to avoid the potential collision between a robot and a human subject and guarantee the safety of human-robot interaction \cite{sajedi2022uncertainty}. On the other hand, the detected human motion can be further used for human motion analysis and prediction to enable robots to comprehend human intention and enhance the efficiency of human-robot interaction \cite{unhelkar2018human}. In this paper, we aim to develop an effective human motion detection algorithm with high accuracy and efficiency.

Detection of moving objects in a video with static background is usually done by separating foreground from background. The moving objects are typically considered as the foreground. Background modeling plays an important role in moving object detection. Many subspace learning methods such as principal component analysis (PCA) have been developed to model background \cite{bouwmans2009subspace} by reducing the dimensionality and learning the intrinsic low-dimensional subspaces. In practice, a background matrix can be generated by concatenating the vectorized versions of background images in a video, which naturally possesses the low-rank structure. Thus sparsity of singular values is pursued for a background matrix. In one of the most popular methods - robust PCA (RPCA) \cite{candes2011robust}, nuclear-norm regularization is used to enforce the matrix low-rankness as a convex relaxation of the matrix rank. Numerous improved variants of RPCA have been proposed for and a comprehensive review can be found \cite{bouwmans2017decomposition}. To further enhance the low-rankness, some other regularizations based on matrix norms, including the matrix max-norm \cite{javed2015background,javed2016spatiotemporal}, have been developed for modeling the background. More recently, adaptive regularization techniques have been developed to promote sparsity and achieve fast convergence of the regularized algorithms. For example, the weighted nuclear norm (WNN) regularization has shown effectiveness in various image and data processing applications \cite{gu2014weighted}, which can be considered as a natural extension of reweighted L1 \cite{candes2008enhancing} and a more general error function based regularization \cite{guo2021novel}. In this paper, we use the WNN-based regularization imposed on the matrix singular values to enforce the adaptive low-rankness. Moreover, a video usually has complicated geometry and varying smoothness in either the spatial domain or the temporal domain. To preserve those geometrical structures in the background, we create a spatial graph and a temporal graph, which are then embedded in the graph regularizations of the background matrix. Generation of the spatial graph is implemented by comparing the patchwise similarity to exploit the nonlocal similarity. To reduce the computational cost, we only consider the $k$-nearest neighboring pixels in terms of similarity when calculating the pairwise similarity. On the other hand, the $\ell_1$-regularization is imposed on the foreground due to its sparsity. Thus far, by integrating various regularizations, we propose a nonconvex spatiotemporal graph regularized moving object detection model, which is solved by applying the alternating direction method of multipliers (ADMM). By introducing a few auxiliary variables and splitting regularizers, we obtain a sequence of subproblems. One quadratic subproblem is solved by gradient descent, and the other subproblems all have closed-form solutions which can be implemented efficiently. Furthermore, we test our algorithm on the two real RGB videos containing a whole-body motion and an arm motion under a static background, respectively. Performance is compared with other related methods in terms of background recovery and foreground detection accuracy.

The rest of this paper is organized as follows. In Section \ref{sec:LR}, we provide a brief introduction of moving object detection and low-rank based models. In Section \ref{sec:method}, we propose a novel spatiotemporal dual graph regularized moving object detection method based on the WNN regularization. Numerical experiments on two realistic videos with moving objects and the results are reported in Section \ref{sec:exp}. Finally, conclusions of this research and future work are presented in Section \ref{sec:con}.

\section{Low-Rank Models}\label{sec:LR}
Throughout the paper, we use boldface lowercase letters to denote vectors and uppercase letters to denote matrices. For $p\geq1$, the $\ell_p$-norm of a vector $\vx\in\R^n$ is given by $\norm{\vx}_p=(\sum_{i=1}^n|x_i|^p)^{1/p}$. The entry-wise $\ell_1$-norm of a matrix $X\in\R^{n\times m}$ is defined as $\norm{X}_1=\sum_{i,j}|x_{ij}|$ where $x_{ij}$ is the $(i,j)$th entry of $X$. The Frobenious norm of $X$, denoted by $\norm{X}_F$, is defined as $\sqrt{\sum_{i,j}|x_{ij}|^2}$. The nuclear norm of $X$, denoted by $\norm{X}_n$, is defined as the sum of all singular values of $X$. We use the symbol $\diag(\vx)$ to denote a diagonal matrix whose diagonal entries form the vector $\vx$, and $I_n$ as the $n$-by-$n$ identity matrix.

Consider a video with static backgrounds consisting of $m$ frames of gray-scale images with size $n_1\times n_2$. By reshaping each image as a vector, we convert a video to a matrix $D$ of size $n\times m$ where $n=n_1n_2$ is the number of total spatial pixels. Assume that $D$ can be decomposed into the background component $L$ and the foreground component $S$, where $L,S\in\R^{n\times c}$. Here we let $S$ correspond to the moving object. That is, we have $D=L+S$ in the noise-free case. In order to retrieve $L$ and $S$ from $D$ simultaneously, we need to apply regularization techniques on both variables. Due to the static background, the matrix $L$ typically has low-rank structures. In the meanwhile, the object occupies a small portion of each frame and thereby $S$ is sparse. Thus we could consider the problem
\[
\min_{L,S}\mathrm{rank}(L)+\lambda\norm{S}_1\quad \st\quad D=L+S.
\]
Here $\lambda>0$ is a regularization parameter and $\rank(L)$ equals the number of nonzero singular values of $L$. Since this problem is NP-hard, matrix rank is replaced by the nuclear norm which leads to the RPCA model \cite{candes2011robust}
\[
\min_{L,S}\norm{L}_n+\lambda\norm{S}_1\quad\st\quad D=L+S.
\]
In some RPCA variants \cite{shen2014online,javed2015background}, the matrix max-norm based regularizer has been used to replace the nuclear norm
\[
\min_{L,S}\norm{L}_{\max}+\lambda\norm{S}_1\quad \st\quad D=L+S.
\]
Here the max-norm of $L$ is given by $\norm{L}_{\max}=\min_{L=UV'}\norm{U}_{2\to\infty}\norm{V}_{2\to\infty}$ where $V'$ is the transpose of $V$ and $\norm{U}_{2\to\infty}=\max_{\norm{\vx}_2=1}\norm{U\vx}_\infty$. See \cite{srebro2005rank} for the connections between the matrix nuclear norm and the max-norm. They both are convex and can be used to describe the low-rankness of the background matrix. 
More recently, adaptive regularizers such as reweighted L1\cite{candes2008enhancing} and its more general version - error function based regularization \cite{guo2021novel} have shown the advantages over the traditional $\ell_1$-regularization in terms of sparsity and convergence speed. They can be naturally extended to the singular values to promote the low-rankness. In this paper, we adopt a weighted nuclear matrix norm \cite{gu2014weighted} as the regularizer
\begin{equation}\label{eqn:matnorm}
\norm{L}_{W,*}:=\sum_{i}w_i\sigma_i(L),
\end{equation}
where $\sigma_i(L)$ is the $i$th singular value of $L$ in the decreasing order and $w_i\geq0$ is the $i$-th weight. Although the weighted nuclear norm minimization (WNNM) is not convex in general, it has shown outstanding performance in a lot of image processing applications \cite{gu2014weighted}. Here we use the exponential function-based weighting scheme that will be detailed in the next section.

\section{Proposed Method}\label{sec:method}
Moving object detection (MOD) is one of the fundamental tasks in robotic applications. The problem is typically cast as the foreground and the background separation. Besides the low-rankness assumption of the background matrix, we can use spatial and temporal graphs to handle the sophisticated geometry. To split multiple regularization terms in the proposed MOD model, we apply the ADMM framework to design an efficient algorithm.

\subsection{Spatial and Temporal Graph Laplacians}\label{sec:lap}
In what follows, we will describe the generation of spatial and temporal graph Laplacians and their corresponding graph regularizers on the background.

For a reshaped video $D\in\mathbb{R}^{n\times m}$, rows and columns of $D$ correspond to the spatial and the temporal samples, respectively. Consider a weighted temporal graph $G_t=(V_t,E_t,A_t)$ where $V_t=\{\vv_i^t\}_{i=1}^m$ is a set of temporal samples, $E_t$ is an edge set and $A_t\in\R^{m\times m}$ is the adjacency matrix which defines the weights. First, we generate the adjacency matrix $A_t$ whose $(i,j)$-th entry is given by
\[
(A_t)_{i,j}=\exp\left(-\frac{\norm{\vv_i^t-\vv_j^t}_2^2}{h_t^2}\right),\quad i,j\in\{1,\ldots,m\}
\]
where $h_t>0$ is a temporal filtering parameter. Let $W_t$ be the degree matrix of $G_t$ where $(W_t)_{i,i}=\sum_{j=1}^m(A_t)_{i,j}$. Next we define the symmetrically normalized temporal graph Laplacian $\Phi_t\in\R^{m\times m}$ given by
\[
\Phi_t=I_m-W_t^{-1/2}A_tW_t^{-1/2}.
\]
Note that $W_t^{-1/2}$ is a diagonal matrix whose $i$-th diagonal entry is $(W_t)_{i,i}^{-1/2}$.

Likewise, we consider a weighted spatial graph $G_s=(V_s,E_s,A_s)$ where $V_s=\{\vv_i^s\}_{i=1}^n$ is a set of spatial samples, $E_s$ is the edge set and $A_s\in\R^{n\times n}$ is a spatial adjacency matrix. Slightly different from the construction of $A_t$, we consider the patchwise similarity in the spatial domain for $A_s$. Specifically, the $(i,j)$-th entry of $A_s$ is given by
\[
(A_s)_{i,j}=\exp\left(-\frac{\norm{\mathcal{N}(\vv_i^s)-\mathcal{N}(\vv_j^s)}_F^2}{h_s^2}\right),\, i,j\in\{1,\ldots,n\}
\]
where $\mathcal{N}(\vv_i^s)\in\R^{p^2\times m}$ is a reshaped version of the video patch centered at the $i$-th pixel and $h_s>0$ is the spatial filtering parameter. To reduce the computational cost, we consider the $k$-nearest neighbors in terms of location for calculating $A_s$. Specifically, we use the four-nearest neighboring spatial pixels to compute the patch-based similarity for generating the spatial adjacency matrix $A_s$. Likewise we use the four-nearest neighboring temporal pixels to compute $A_t$.  Moreover, it is worth noting that Gaussian smoothing could be embedded to the calculation of patchwise similarity in the presence of noise. Now we define the symmetrically normalized graph Laplacian in the spatial domain as
\[
\Phi_s=I_n-W_s^{-1/2}A_sW_s^{-1/2}.
\]
Similar to $W_t$, $W_s$ is the degree matrix corresponding to $G_s$ which can be obtained using $A_s$. Furthermore, we save all graph Laplacians as sparse matrices to circumvent the out-of-memory issue.

\subsection{Robust Dual-Graph Regularized Method}
Let $D\in\R^{n\times m}$ be the reshaped video with $n$ pixels and $m$ frames. Assume that $\Phi_s\in\R^{n\times n}$ and $\Phi_t\in\R^{m\times m}$ are the graph Laplacians in the respective spatial and temporal domains, which are obtained from Section~\ref{sec:lap}. We propose a robust foreground-background separation model of the form
\[\begin{aligned}
&\min_{L,S\in\R^{n\times m}}\norm{D-L-S}_1+\lambda_1\norm{L}_{W,*}+\lambda_2
\norm{S}_1\\
&+\frac{\gamma_1}2\tr(L^T\Phi_s L)+\frac{\gamma_2}2\tr(L\Phi_t L^T).
\end{aligned}\]
Here we adopt the $L_1$-norm in the first data fidelity term to enforce the robustness of the method and suppress the outliers for recovering the low-rank component. The last two terms are the graph regularizations in the respective spatial and the temporal domains, which can be used to enforce the spatiotemporal smoothness for the background. By introducing an auxiliary variables $U$ and $V$, we rewrite the above problem
\begin{multline*}
\min_{L,S,U,V}\norm{V}_1+\lambda_1\norm{U}_{W,*}+\lambda_2
\norm{S}_1+\frac{\gamma_1}2\tr(L^T\Phi_s L)\\
+\frac{\gamma_2}2\tr(L\Phi_t L^T),\quad \st \quad U=L,\,D-L-S=V.
\end{multline*}
Define the augmented Lagrangian
\begin{multline*}
\mathcal{L}
=\norm{V}_1+\lambda_1\norm{U}_{W,*}+\lambda_2
\norm{S}_1+\frac{\gamma_1}2\tr(L^T\Phi_s L)\\
+\frac{\gamma_2}{2}\tr(L\Phi_t L^T)+\frac{\rho_1}2\norm{U-L+\widetilde{U}}_F^2+\frac{\rho_2}2\norm{D-L-S+V+\widetilde{V}}_F^2.
\end{multline*}
Based on the ADMM framework, we derive the algorithm
\[
\left\{\begin{aligned}
L&\leftarrow \argmin_L\frac{\gamma_1}2\tr(L^T\Phi_s L)+\frac{\gamma_2}2\tr(L\Phi_t L^T)\\&
+\frac{\rho_1}2\norm{U-L+\widetilde{U}}_F^2+\frac{\rho_2}2\norm{D-L-S+V+\widetilde{V}}_F^2\\
S&\leftarrow \argmin_S\lambda_2\norm{S}_1+\frac12\norm{D-L-S}_F^2\\
U&\leftarrow \argmin_U\lambda_1\norm{U}_{W,*}+\frac{\rho_1}2\norm{U-L+\widetilde{U}}_F^2\\
&=\argmin_U\frac{\lambda_1}{\rho_1}\norm{U}_{W,*}+\frac12\norm{U-L+\widetilde{U}}_F^2\\
V&\leftarrow\argmin_V\norm{V}_1+\frac{\rho_2}2\norm{D-L-S+V+\widetilde{V}}_F^2\\
\widetilde{U}&\leftarrow \widetilde{U}+(U-L)\\
\widetilde{V}&\leftarrow \widetilde{V}+(D-L-S+V)
\end{aligned}\right.
\]
For the first $L$-subproblem, we apply the gradient descent to solve it. 
Note that although the critical equation is a Sylvester equation, the giant matrix $\Phi_s$ will make the standard Sylvester solver very slow.
The gradient of the objective function is
\[\begin{aligned}
\nabla f(L)&=\gamma_1\Phi_sL+\gamma_2 L\Phi_t+\rho_1(L-U-\widetilde{U})\\
&+\rho_2(L-D+S-V-\widetilde{V}).
\end{aligned}
\]
Note that $\frac{d}{dX}\tr(X^TAX)=(A+A^T)X=2AX$ if $A$ is a symmetric matrix.
At each inner loop, we update $L$ with fixed $S,U,\widetilde{U},V,\widetilde{V}$ via
\begin{equation}\label{eqn:Lupdate}
L\leftarrow L-dt\cdot \nabla f(L),
\end{equation}
where $dt>0$ is a step size. It can be empirically shown that only a few steps of gradient descent is required here.
The $S$-subproblem has the closed-form solution
\begin{equation}\label{eqn:Supdate}
S\leftarrow \shrink(D-L, \lambda_2).
\end{equation}
Here the shrinkage operator is defined as $(\shrink(A,\mu))_{ij}=\mathrm{sign}(a_{ij})\cdot\max\{|a_{ij}|-\mu,0\}$ where $a_{ij}$ is the $(i,j)$-th entry of $A$. It can be shown that the $U$-subproblem has the closed-form solution via a weighted version of the singular value thresholding operator (SVT)
\begin{equation}\label{eqn:Uupdate}
U\leftarrow A\widetilde{\Sigma}B,\quad \widetilde{\Sigma}=\diag(\shrink(\sigma(\widehat{L}),w_i\lambda_1/\rho_1))
\end{equation}
where $A\Sigma B$ is the singular value decomposition (SVD) form of the matrix $\widehat{L}:=(L-\widetilde{U})$, $\sigma(\widehat{L})$ is the vector containing all the singular values of $\widehat{L}$ with $\sigma_i(\widehat{L})$ as its $i$-th component. Here the weights are constructed iteratively based on the singular values of the matrix $L$ from the previous iteration:
\begin{equation}\label{eqn:Wupdate}
w_i=\exp(-\sigma_i^2(\widehat{L})/{\sigma^2}).
\end{equation}

Finally, the $V$-subproblem is similar to the $S$-subproblem with the closed-form solution and thereby $V$ is updated via
\begin{equation}\label{eqn:Vupdate}
V\leftarrow \shrink(L+S-D-\widehat{V},\rho_2).
\end{equation}

We remove motionless frames in the data set if two consecutive frames have small overall changes, i.e., the $\ell_1$-norm of the difference vector of the two adjacent columns of $D$ is below a threshold. The stopping criteria are based on the relative changes in $L$ and $S$, i.e.,
$
\frac{\norm{L^{i+1}-L^i}_F}{\norm{L^i}_F}<tol
$
and
$
\frac{\norm{S^{i+1}-S^i}_F}{\norm{S^i}_F}<tol
$
where $L^t$ and $S^t$ are the obtained background and foreground matrices at the $i$-th iteration and $tol$ is the preassigned tolerance.
The entire algorithm is summarized in Algorithm~\ref{alg}, which can be extended to handle RGB data sets channelwise and then fuse the results from various color channels. In this work, we focus on gray scale videos by converting all RGB data to gray scale ones.

\begin{algorithm}
\caption{Robust Dual-Graph Regularized Moving Object Detection}\label{alg}
\begin{algorithmic}
\State\textbf{Inputs}: reshaped test video $D\in\R^{n\times m}$, graph filtering parameters $h_s,h_t>0$, parameters $\lambda_1,\lambda_2,\gamma_1,\gamma_2,\rho_1,\rho_2>0$, maximum outer loops $T_{out}$, maximum inner loops $T_{in}$, tolerance $tol$
\State\textbf{Outputs}: background $L$ and foreground $S$
\State Generate graph Laplacians $\Phi_t$ and $\Phi_s$
\State Initialize $L$ and $S$
\For{$i=1,2,\ldots,T_{out}$}
        \For{$j=1,2,\ldots,T_{in}$}
        \State Update $L$ via \eqref{eqn:Lupdate}
        \EndFor
        \State Update $S$ via \eqref{eqn:Supdate}
        \State Update $U$ via \eqref{eqn:Uupdate} and singular values $\sigma_i(\widehat{L})$
        \State Update $W$ via \eqref{eqn:Wupdate}
        \State Update $V$ via \eqref{eqn:Vupdate}
        \State $\widetilde{U}\leftarrow \widetilde{U}+(U-L)$
        \State $\widetilde{V}\leftarrow \widetilde{V}+(D-L-S+V)$
        \State {Exit the loop if the stopping criteria are met.}
\EndFor
\end{algorithmic}
\end{algorithm}

\section{Numerical Experiments}\label{sec:exp}
In this section, we will test the proposed Algorithm~\ref{alg} on two simulated moving object images. For comparison, we include three closely related algorithms based on the fast robust principal component analysis (RPCA) \cite{candes2011robust}: (1) Largangian optimization method for unconstrained RPCA (LAGO) (2) stable principal component pursuit (SPCP) \cite{driggs2019adapting} and (3) SPGL1 \cite{van2009probing} for solving the problem $\min_{L,S}\max\{\norm{L}_*,\lambda\norm{S}_1\}$ subject to $\norm{D-L-S}_F\leq\varepsilon$. Their source codes can be found in fastRPCA \url{https://github.com/stephenbeckr/fastRPCA} \cite{aravkin2014}. There are two groups of metrics for comparing the performance, i.e., comparing the foreground and the background. First, the static background image is extracted from the low-rank component of the given video. We take the mean column of the low-rank matrix $L$ and then reshape it as a matrix. We use the following metrics to evaluate the background recovery quality:
\begin{itemize}
\item relative error (RE): $\mathrm{RE}(\widehat{L},L)=\frac{\norm{L-\widehat{L}}_F}{\norm{L}_F}$;
\item peak signal-to-noise ratio (PSNR): $\mathrm{PSNR}(\widehat{L},L)=20\log({I_{\max}}/\sqrt{\norm{\widehat{L}-L}_F^2/(n_1n_2)})$.
\end{itemize}
Here $\widehat{L}$ is the estimate of the ground truth $L\in\R^{n_1\times n_2}$, and $I_{\max}$ is the maximum image intensity set as 1. In our experiments, all of the videos to be processed are scaled to the range $[0,1]$.

For the foreground assessment, we apply the hard thresholding to extract the foreground masks and then compute the following metrics. Here ground truth foreground masks are manually made. Let $TP$ be the true positive counting the foreground pixels correctly labeled as foreground, $FP$ be the false positive counting the background pixels incorrectly labeled as foreground, and $FN$ be the false negative counting the foreground pixels incorrectly labeled as background. The three metrics are defined as follows.
\begin{itemize}
    \item Precision (Pr): $\mbox{Pr}={TP}/{(TP+FP)}$
    \item Recall (Re):
    $\mbox{Re}={TP}{(TP+FN)}$
    \item F-measure (Fm):
    $\mbox{Fm}=2{\mbox{Re}}/{\mbox{Pr}}$
\end{itemize}
All the three metrics are between 0 and 1. The higher the value is, the more accurate the result is. We also find that various hard thresholding strategies may cause different one or two metrics high while the remaining ones are low.

A Microsoft Azure Kinect Sensor was used to record human motion, including one 1-MP depth sensor, one 7-microphone array, one 12-MP RGB video camera, and one accelerometer and gyroscope (IMU) sensor. Designed to pull together multiple AI sensors in a single device, Azure Kinect sensors have been employed for various applications, such as building telerehabilitation solutions, democratizing home fitness, etc. 
In this study, only the RGB video camera was used to record human motion and test the proposed algorithm. All numerical experiments were run in Matlab R2021a on a desktop computer with Intel CPU i9-9960X RAM 64GB and GPU Dual Nvidia Quadro RTX5000 with Windows 10 Pro.

\subsection{Experiment 1: Whole Body Movement Video}
\begin{figure*}[h]
\centering\setlength{\tabcolsep}{2pt}
\begin{tabular}{ccccc}
\includegraphics[width=0.19\textwidth]{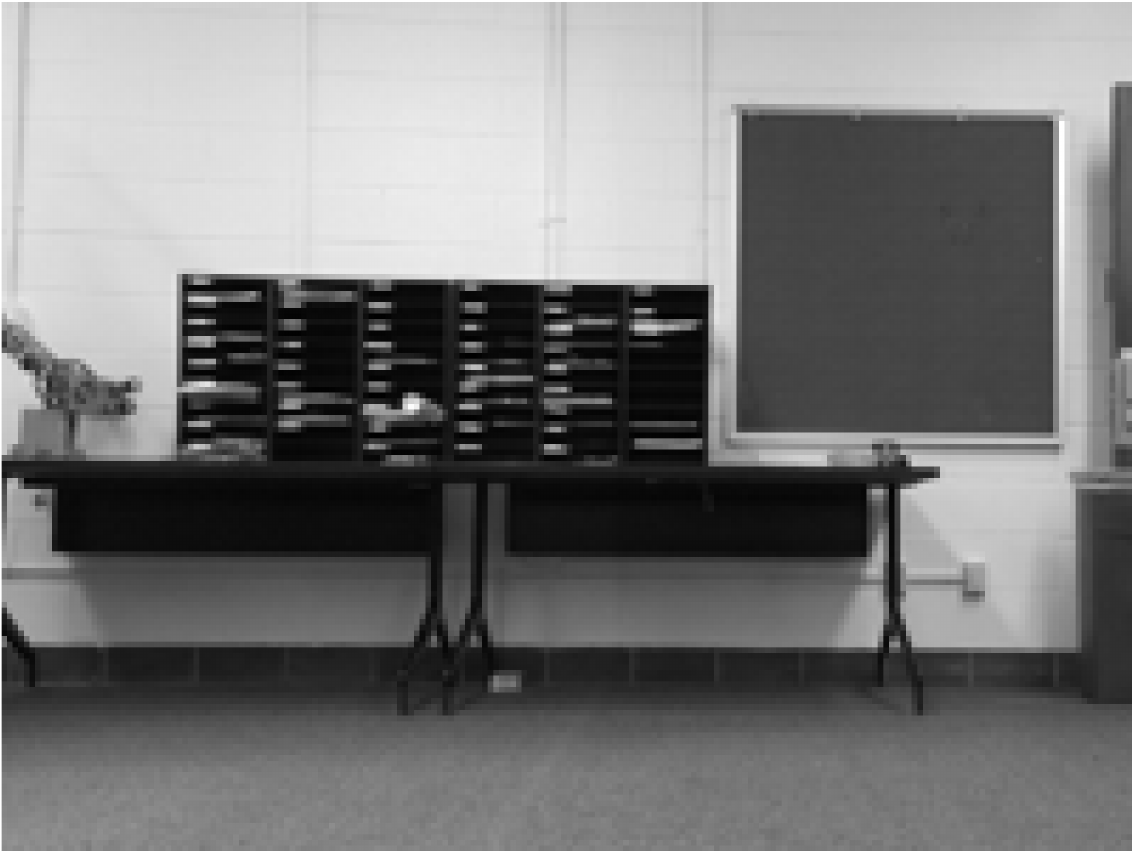}&
\includegraphics[width=0.19\textwidth]{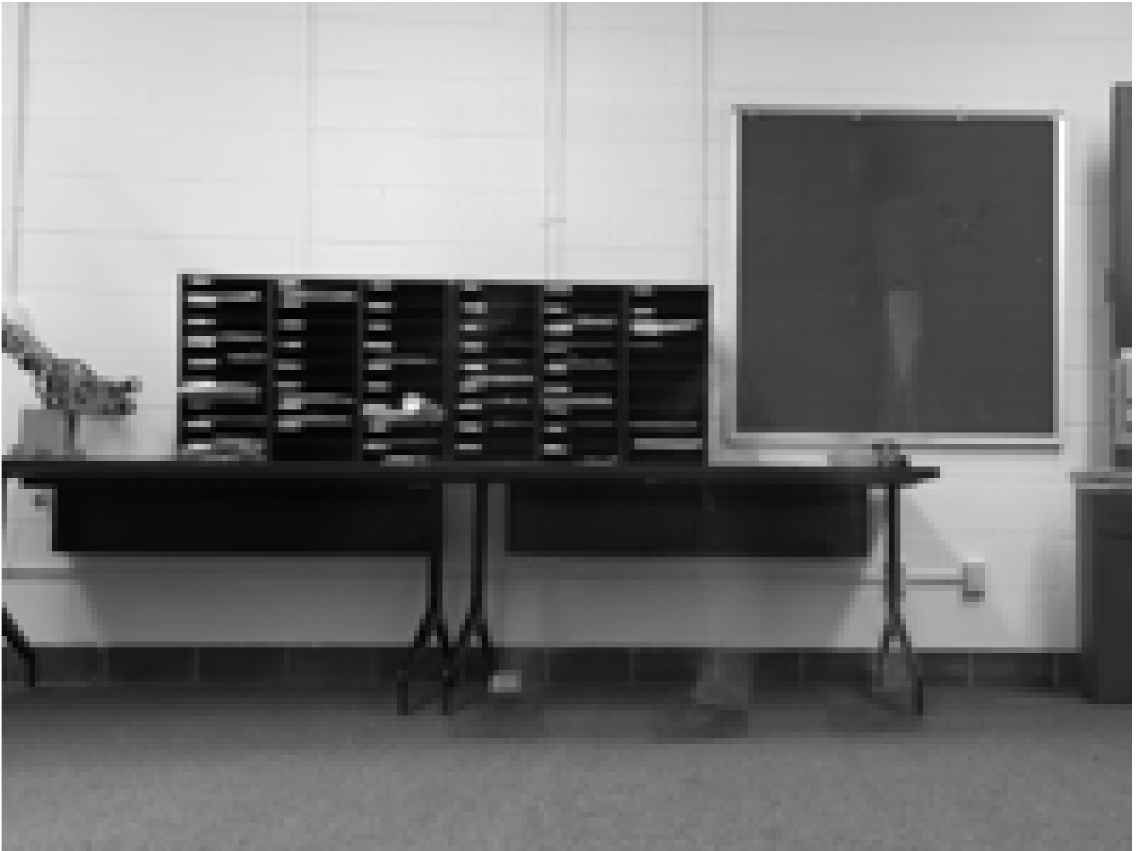}&
\includegraphics[width=0.19\textwidth]{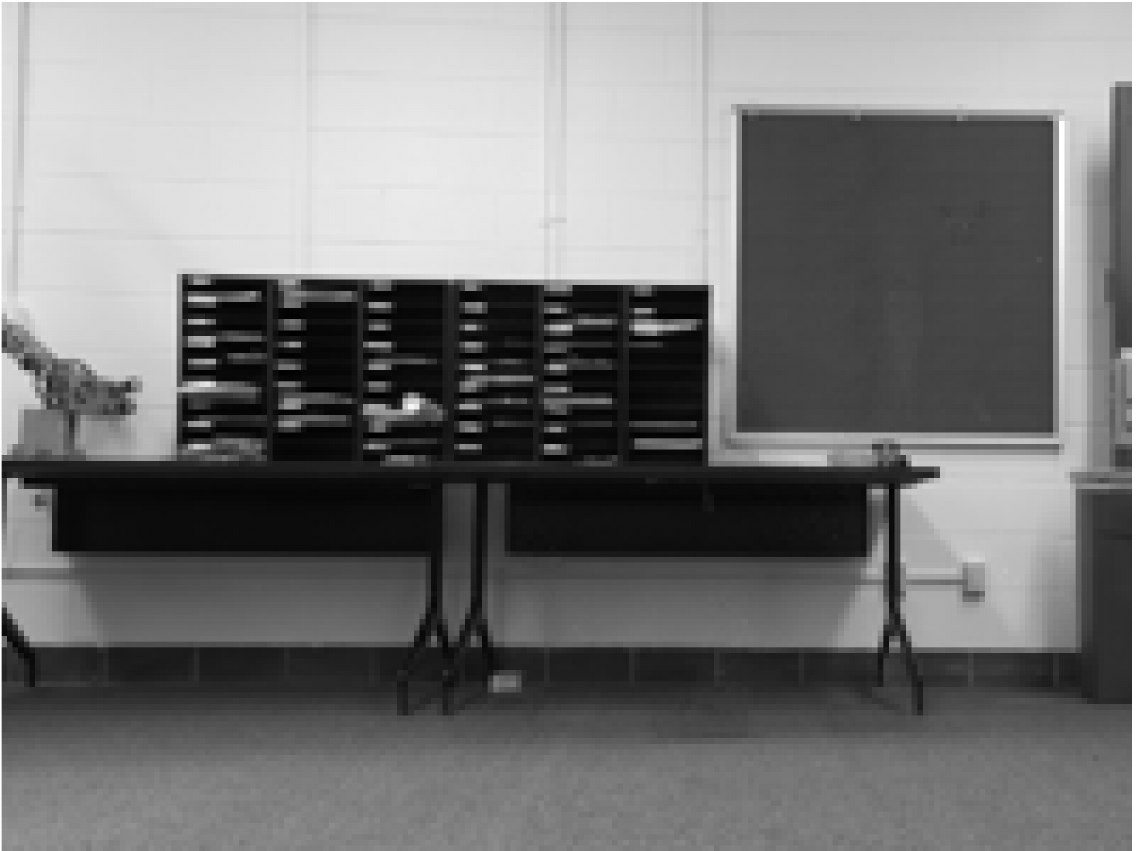}&
\includegraphics[width=0.19\textwidth]{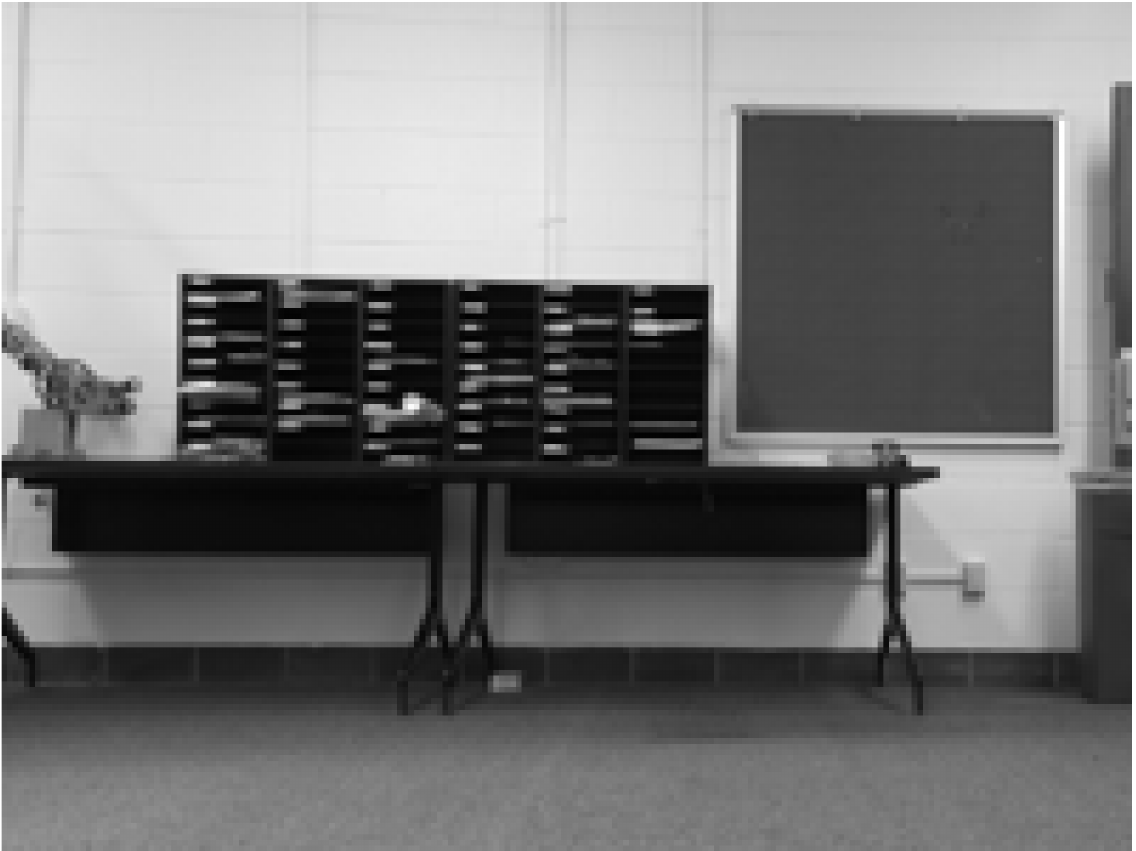}&
\includegraphics[width=0.19\textwidth]{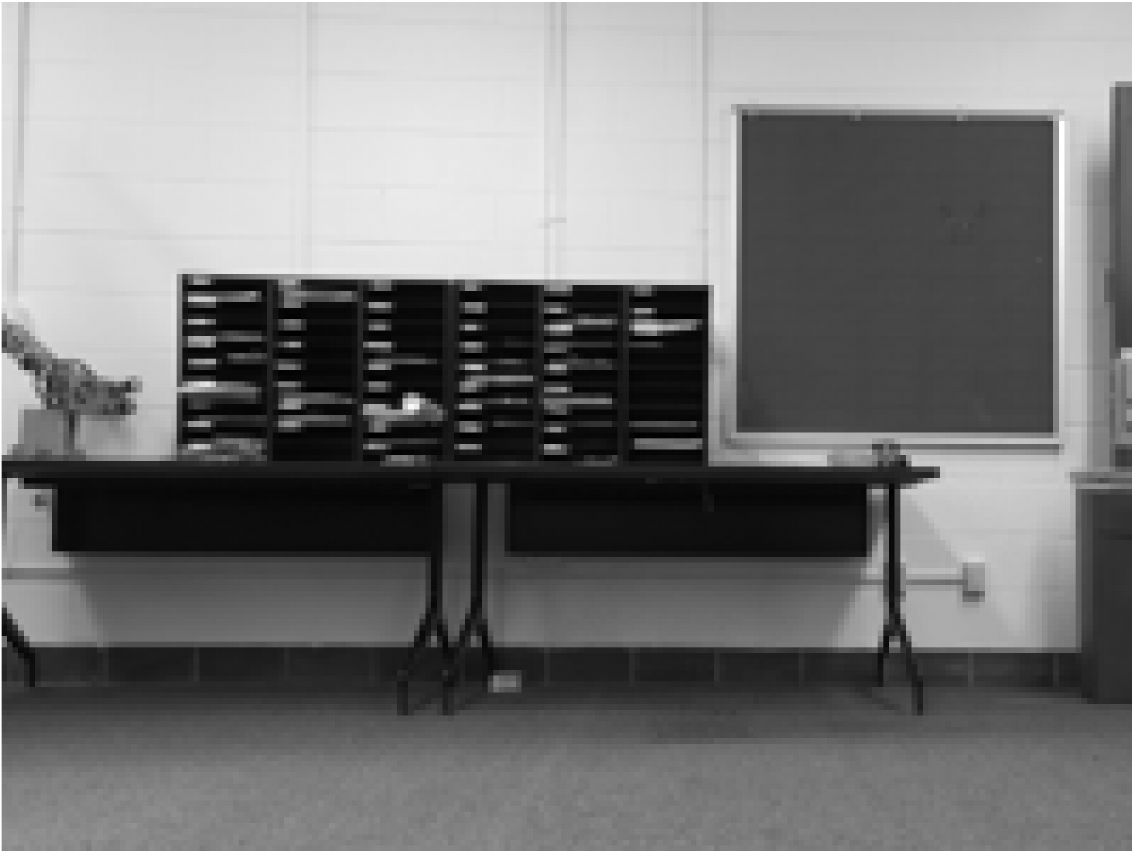}\\[-2pt]
 Ground truth &  LAGO &  SPCP &  SPGL1 &  Alg.1
\end{tabular}
\vspace{-2pt}
\caption{Visualization results of various methods on background recovery for the walking video. }\label{fig:exp1bg}
\end{figure*}

\begin{figure*}[h]
\centering\setlength{\tabcolsep}{2pt}
\begin{tabular}{ccccc}
\includegraphics[width=0.19\textwidth]{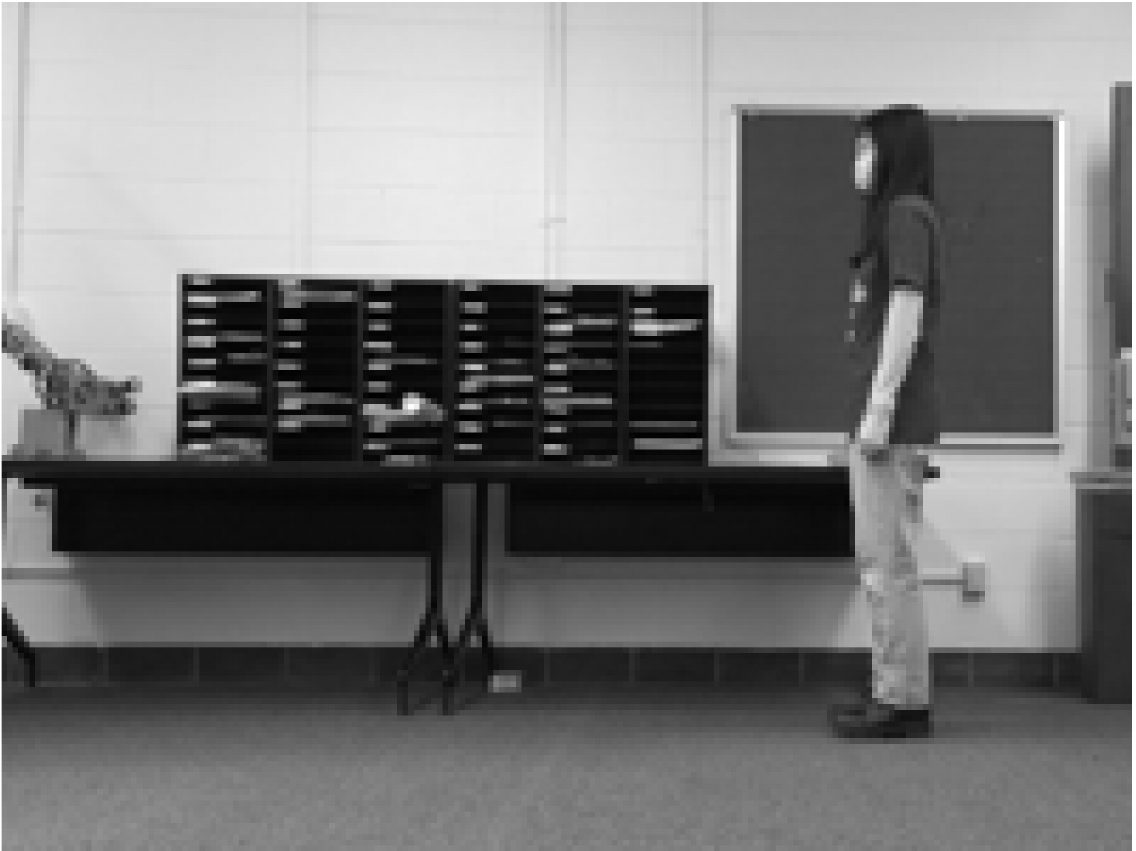}&
\includegraphics[width=0.19\textwidth]{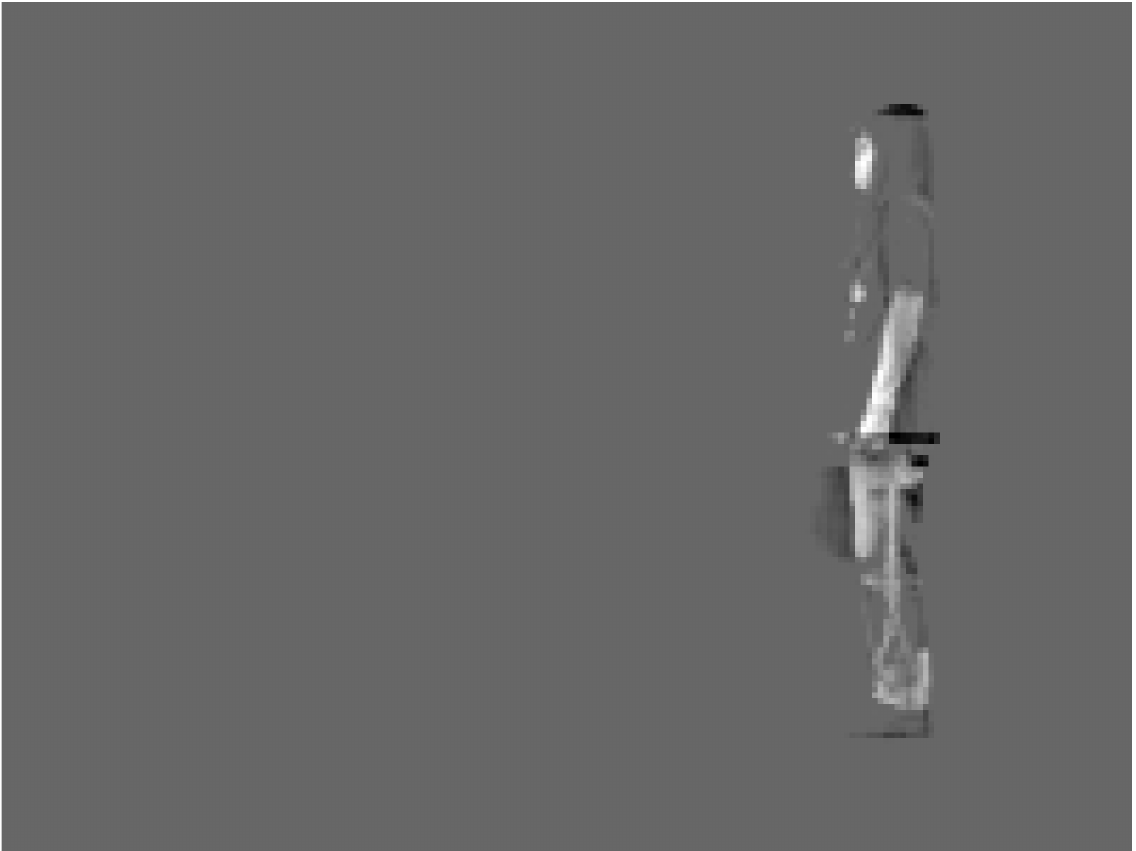}&
\includegraphics[width=0.19\textwidth]{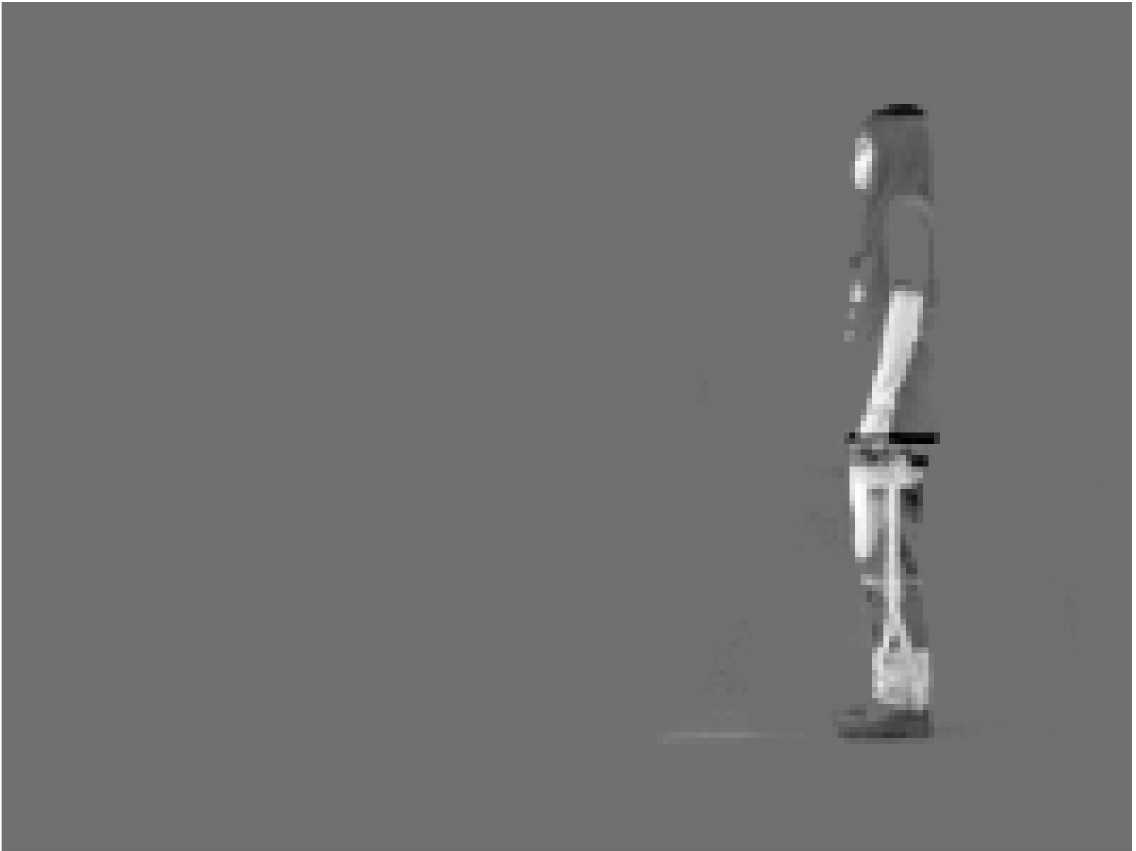}&
\includegraphics[width=0.19\textwidth]{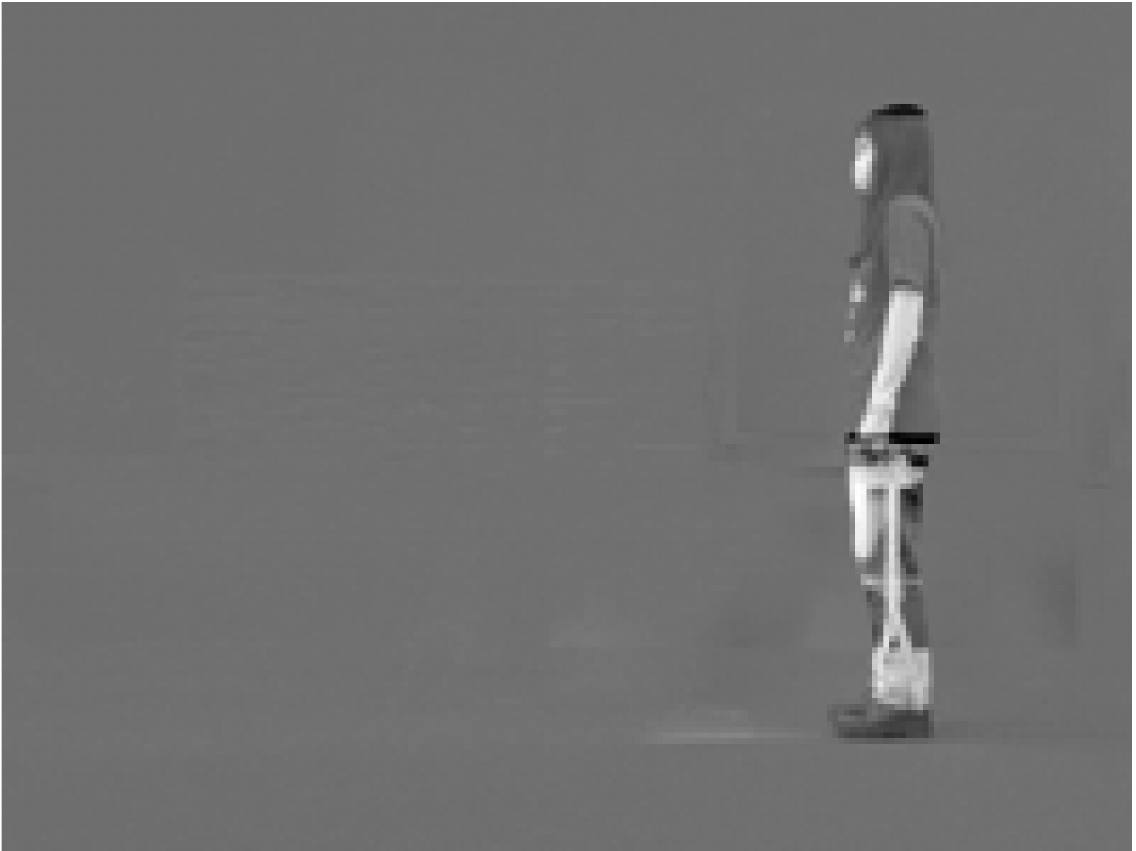}&
\includegraphics[width=0.19\textwidth]{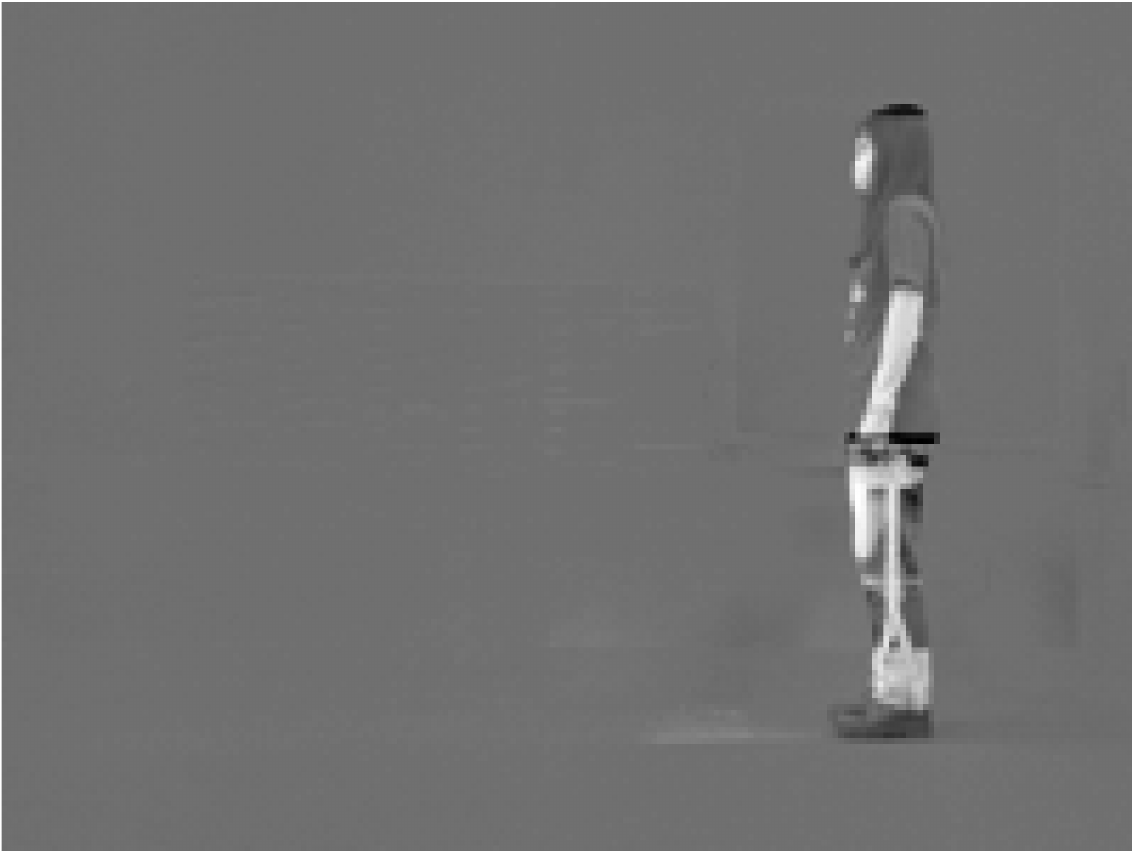}\\
\includegraphics[width=0.19\textwidth]{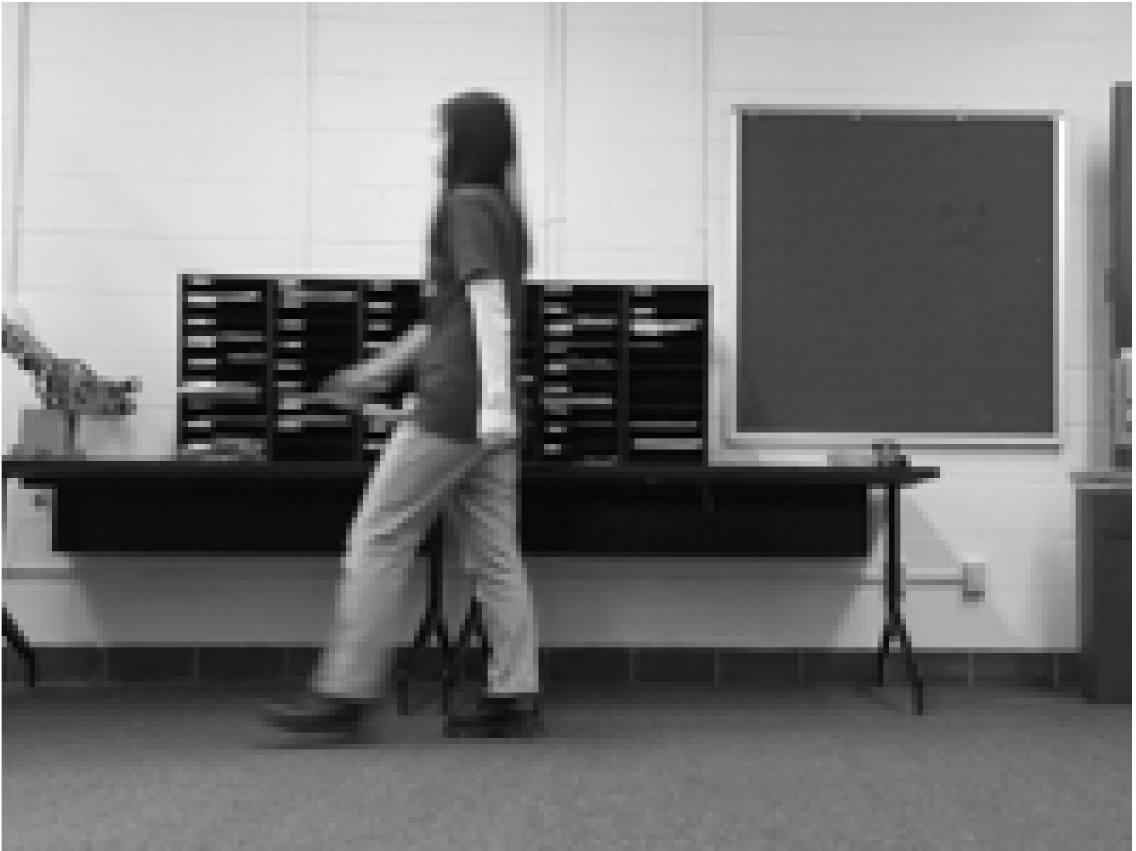}&
\includegraphics[width=0.19\textwidth]{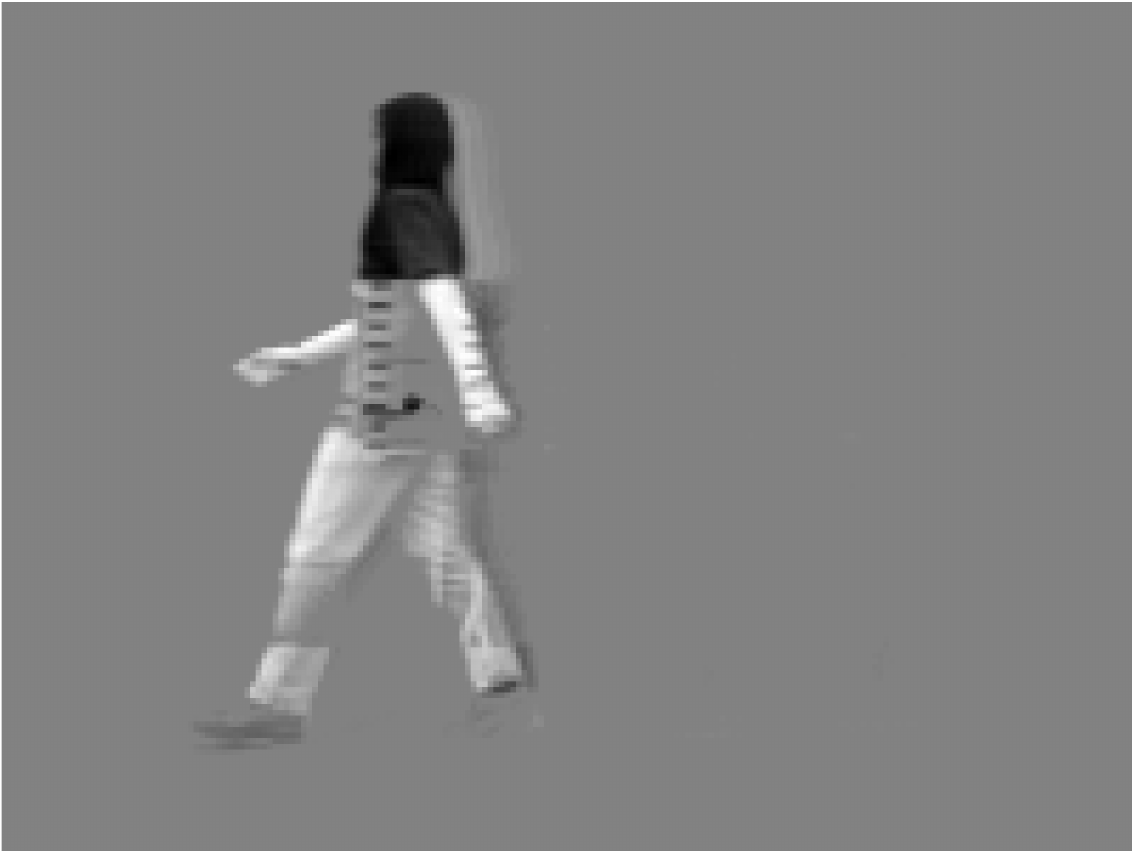}&
\includegraphics[width=0.19\textwidth]{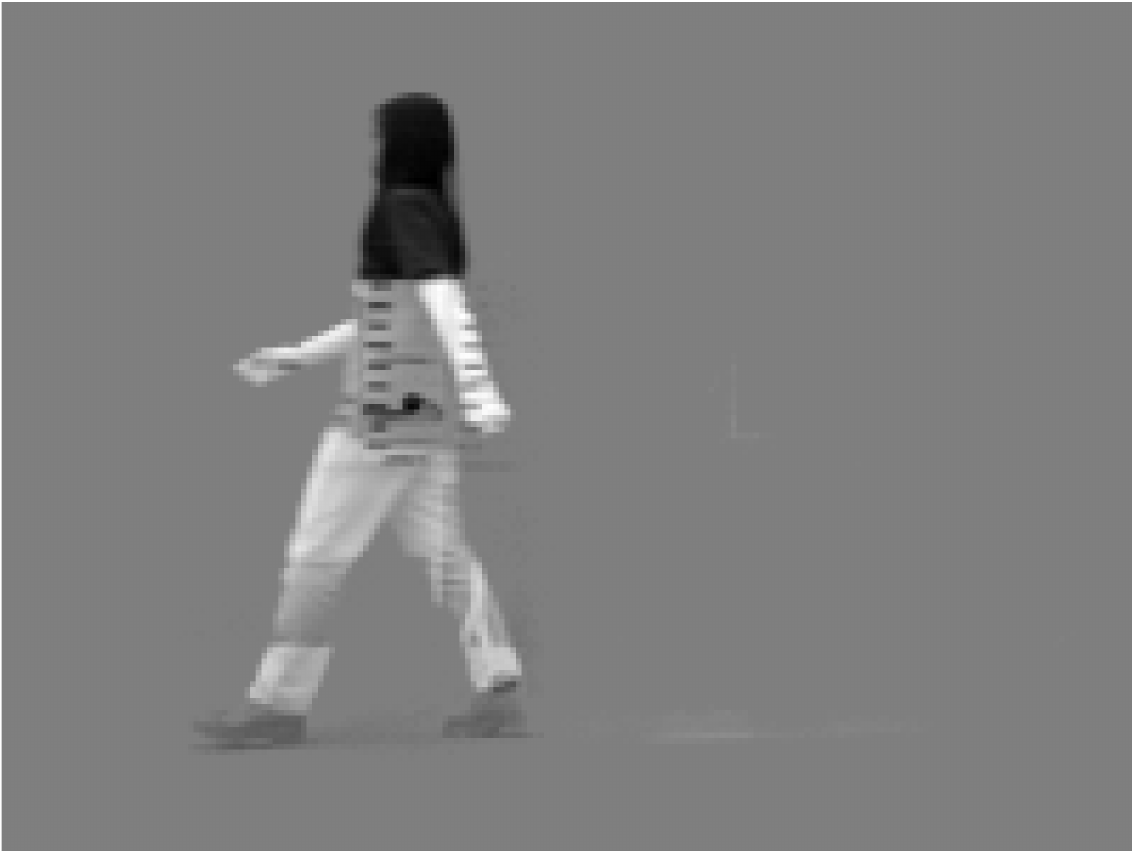}&
\includegraphics[width=0.19\textwidth]{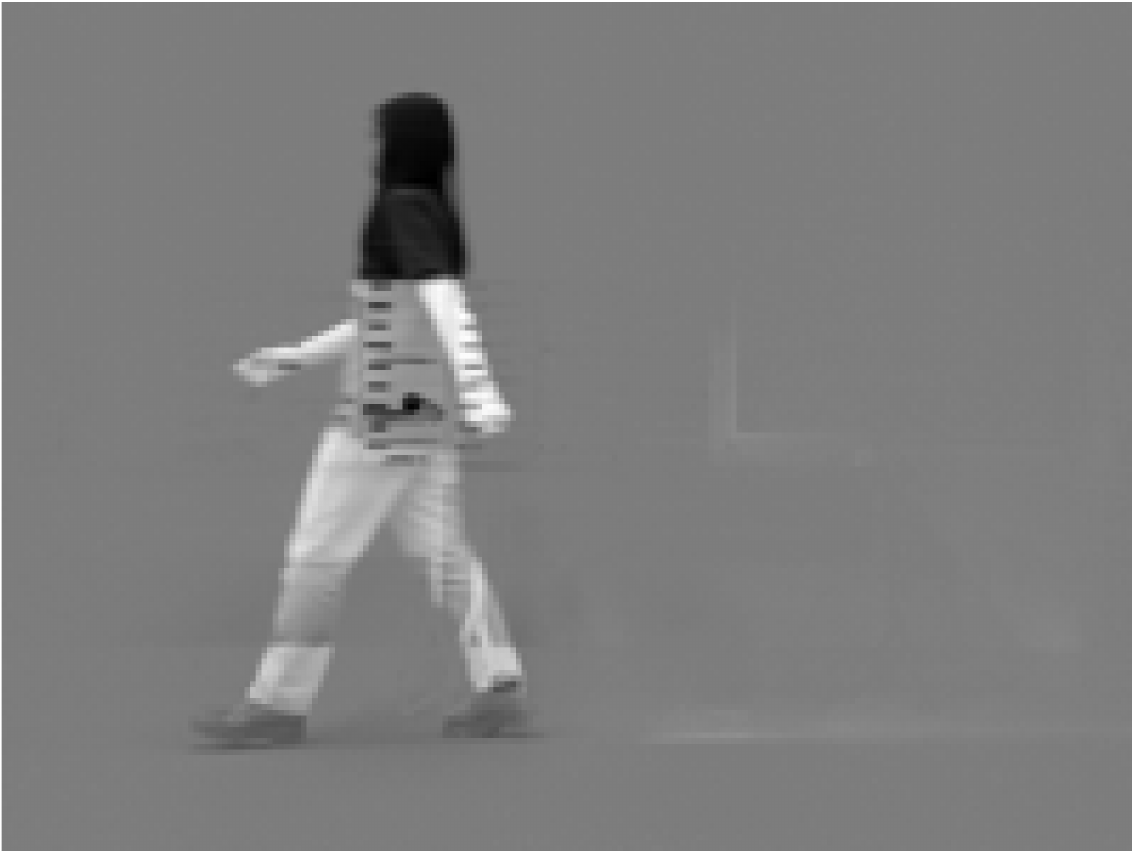}&
\includegraphics[width=0.19\textwidth]{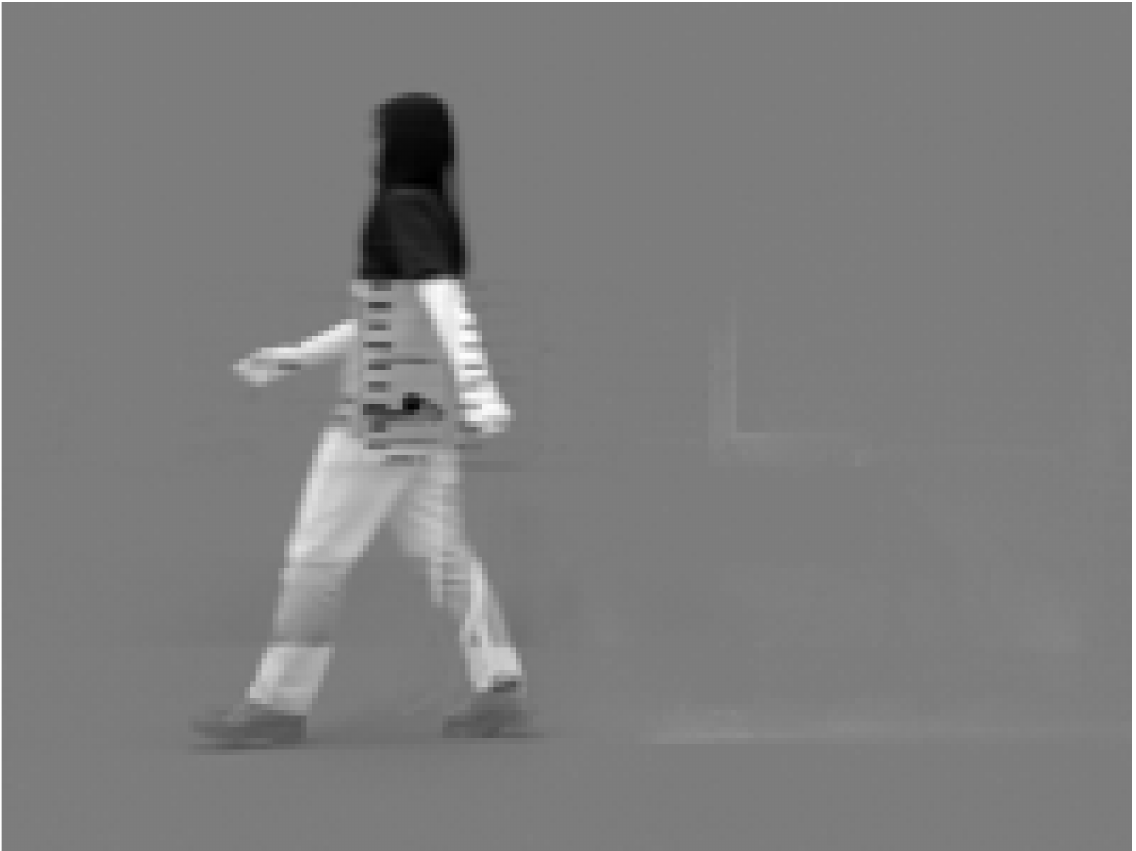}\\[-2pt]
 Original frame &  LAGO &  SPCP &  SPGL1 &  Alg.1
\end{tabular}
\vspace{-2pt}
\caption{Foreground detection for the walking video. The two rows correspond to the first and the last video frames, respectively.}\label{fig:exp1fg}
\end{figure*}

For the first experiment, we consider a video capturing whole-body movement, which was recorded when one student volunteer was walking naturally at an average speed in a lab room. The video of interest consists of 60 frames where each frame has $150\times 200$ pixels. Due to the limited lighting conditions, there are inevitable shadows of the person and brightness variations in the foreground. In Fig.~\ref{fig:exp1bg}, we compare the recovered background from the various methods. For each method, we take the mean column of the recovered $L$ followed by reshaping it as a matrix, i.e., we use the mean of the obtained backgrounds over 60 frames. There are some white spots in the blackboard mistakenly recognized as foreground in the LAGO result and quite a few still exist in the SPCP result. Both SPGL1 and our results can recover the background well except the light shadow on the ground. In Fig.~\ref{fig:exp1fg}, we show the recovered foregrounds at the first and the last frame. The LAGO result has blurry edges for the human body, and both SPGL1 and our results can detect the shadow motion. The quantitative comparison of the recovered foreground and background for all methods is reported in Table~\ref{tab1}. Our method performs best in terms of all the comparison metrics for this video data.

\begin{table}[h]
\centering
\caption{Quantitative comparison for the walking video}\label{tab1}
\vspace{-4pt}
\begin{tabular}{c|cc|ccc}
\hline \hline
    & RE & PSNR & Pr & Re & Fm\\ \hline
LAGO &0.0377&33.67&0.9796&0.4673&0.6328\\
SPCP& 0.0182&39.99&0.9777&0.6354&0.7703 \\
SPGL1 & 0.0148 & 41.81 & 0.9682 & 0.7180&0.8246 \\
Alg.~1 & 0.0145 & 41.95 & 0.9688 & 0.7187&0.8252\\
\hline\hline
\end{tabular}
\end{table}

\subsection{Experiment 2: Arm Movement Video}
\begin{figure*}[h]
\centering\setlength{\tabcolsep}{2pt}
\begin{tabular}{ccccc}
\includegraphics[width=0.19\textwidth]{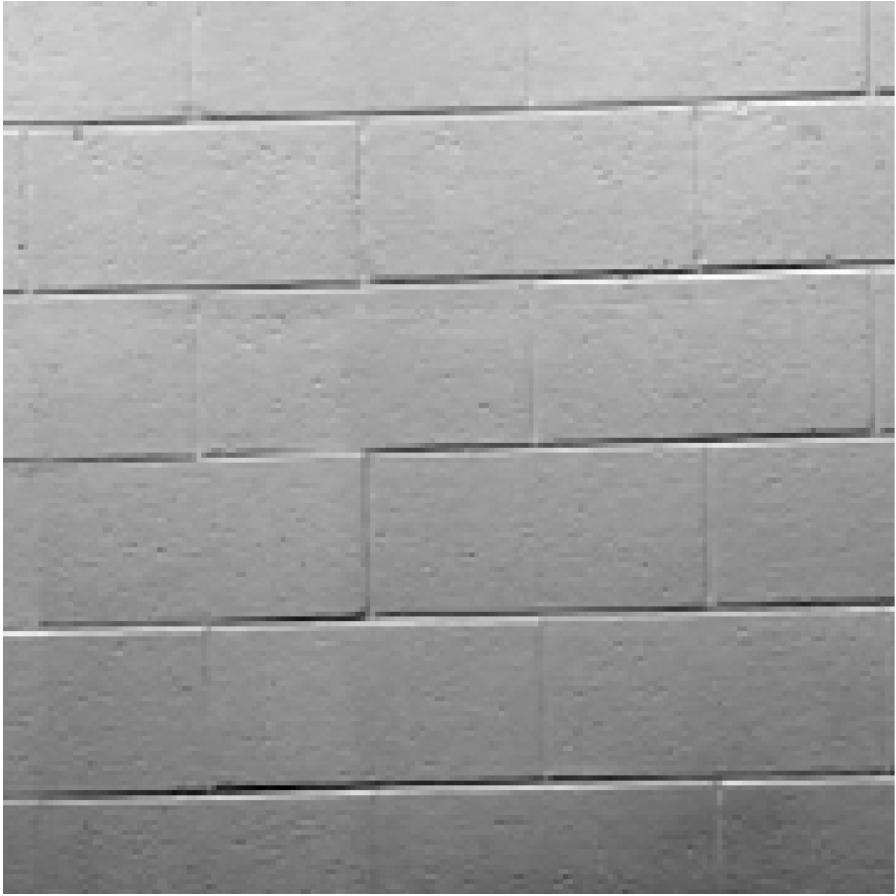}&
\includegraphics[width=0.19\textwidth]{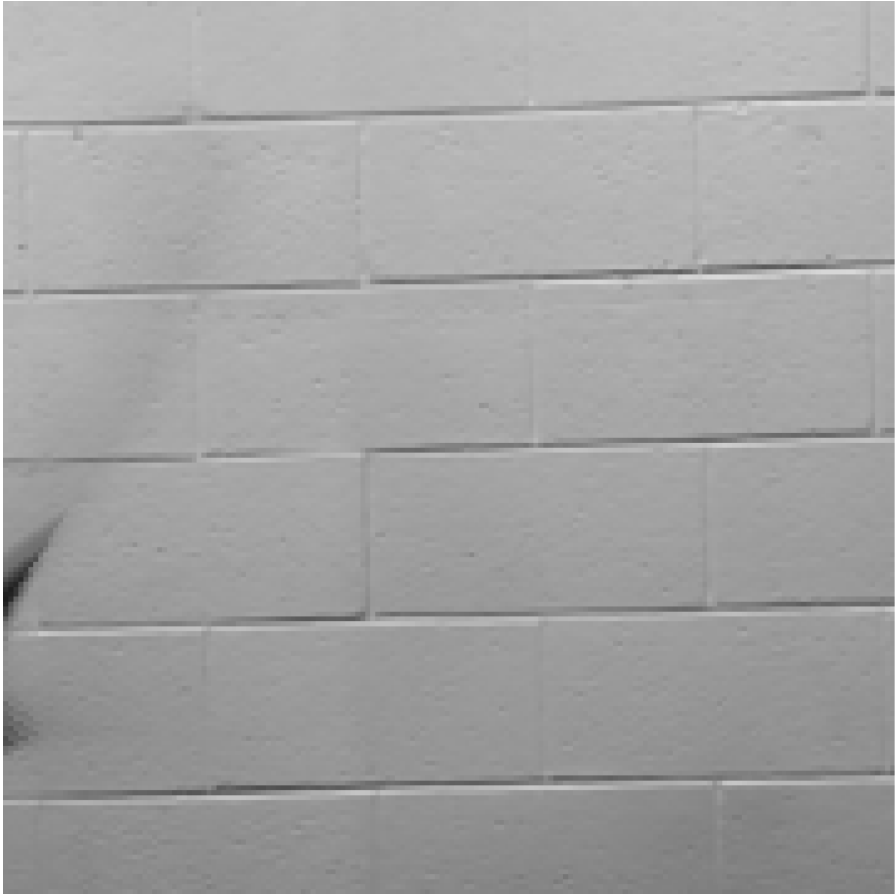}&
\includegraphics[width=0.19\textwidth]{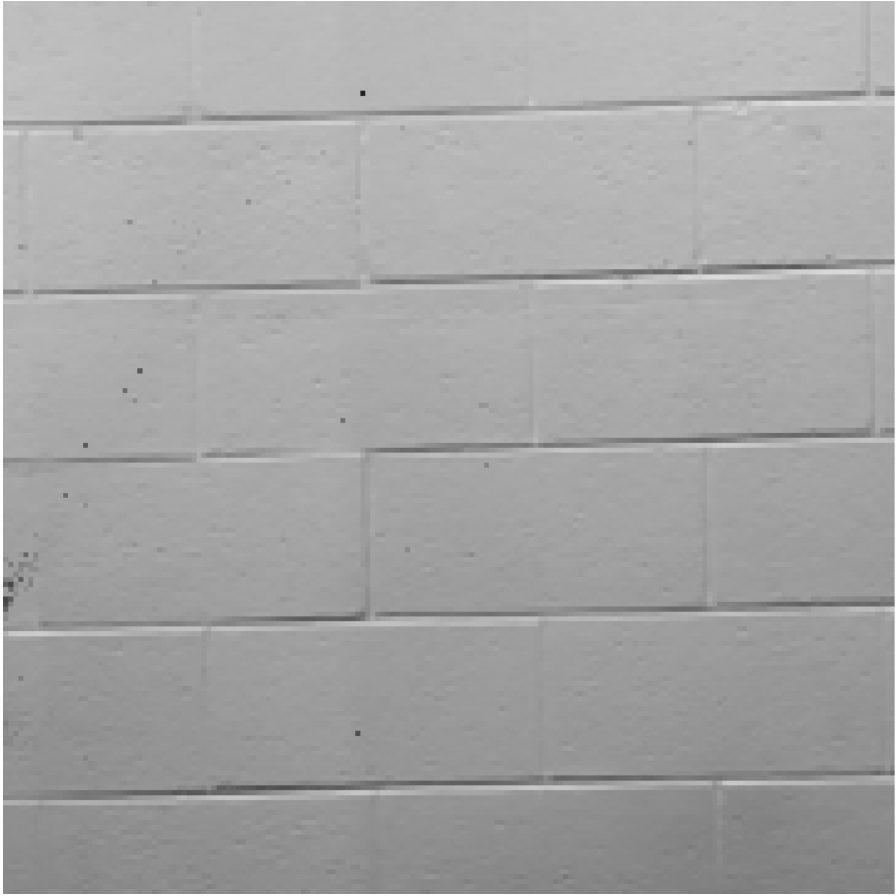}&
\includegraphics[width=0.19\textwidth]{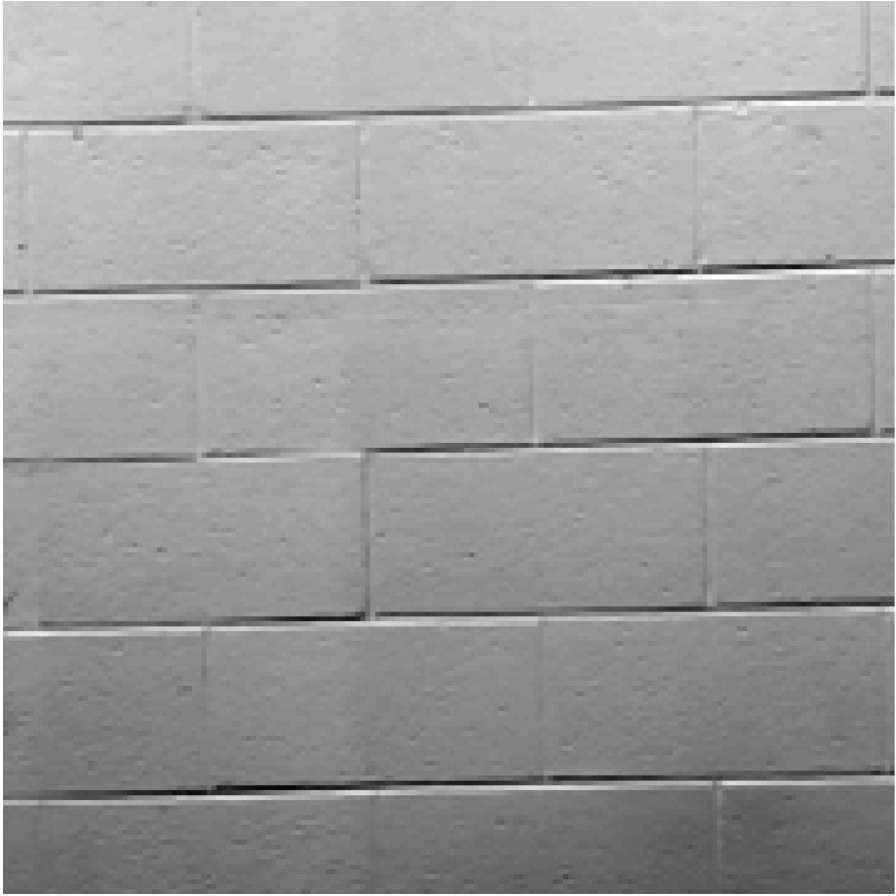}&
\includegraphics[width=0.19\textwidth]{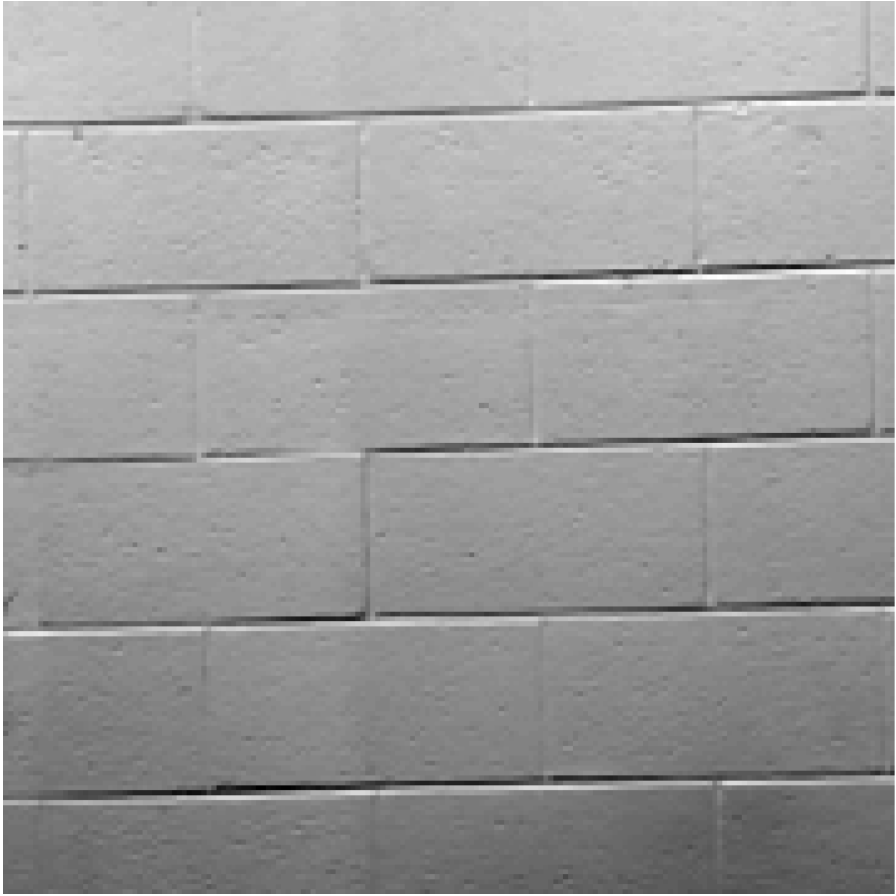}\\[-2pt]
 Ground truth &  LAGO &  SPCP &  SPGL1 &  Alg.1
\end{tabular}
\vspace{-2pt}
\caption{Visualization results of various methods on background recovery for the arm motion video. }\label{fig:exp2bg}
\end{figure*}

In the second experiment, an arm movement video was recorded when the student volunteer rotated her forearm and hand around the elbow joint slowly.
The tested video is generated by removing motionless frames and cropping the region of interest, which consists of 32 frames and each frame has $180\times 180$ pixels. The visual comparison of foreground and background for all results are shown in Fig.~\ref{fig:exp2bg} and Fig.~\ref{fig:exp1fg}, respectively. In Table~\ref{tab2}, we compare the qualities of the recovered background and foreground. Notice that there is movement still left on the left of the LAGO background and foreground results while some speckle noise exist in the SPCP foreground. Both SPGL1 and our approach can separate the foreground and the background clearly. 

In terms of running time, SPCP takes the minimum running time (~0.1 s) while SPGL1 based on Newton's iteration takes about 50 seconds. Both LAGO and our algorithm run about 5 seconds and the graph Laplacian construction can be fast using a small number of neighbors. Overall, our method can keep a good balance in running time and detection accuracy. This phenomenon also applies to the first experiment.

\begin{figure*}[h]
\centering\setlength{\tabcolsep}{2pt}
\begin{tabular}{ccccc}
\includegraphics[width=0.19\textwidth]{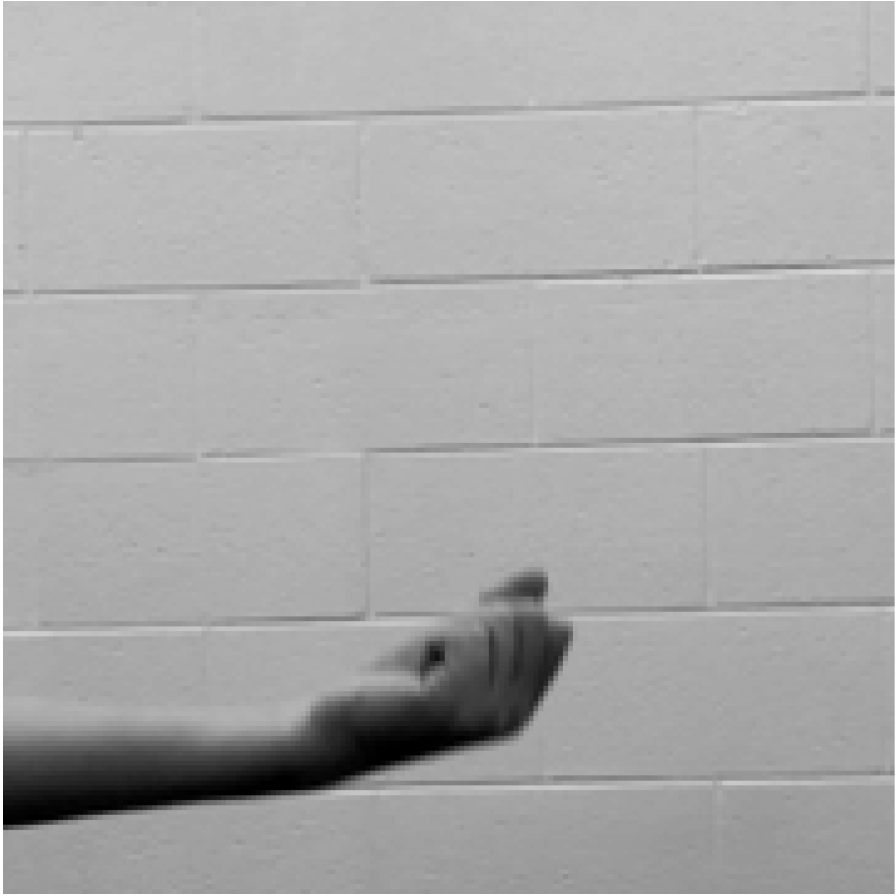}&
\includegraphics[width=0.19\textwidth]{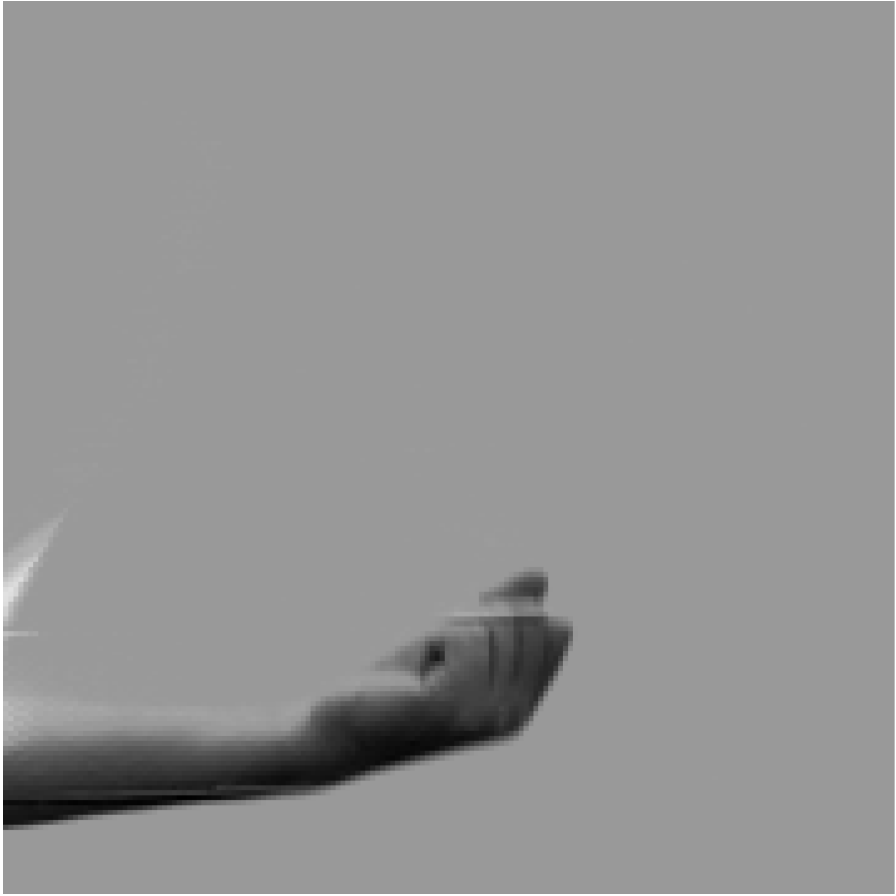}&
\includegraphics[width=0.19\textwidth]{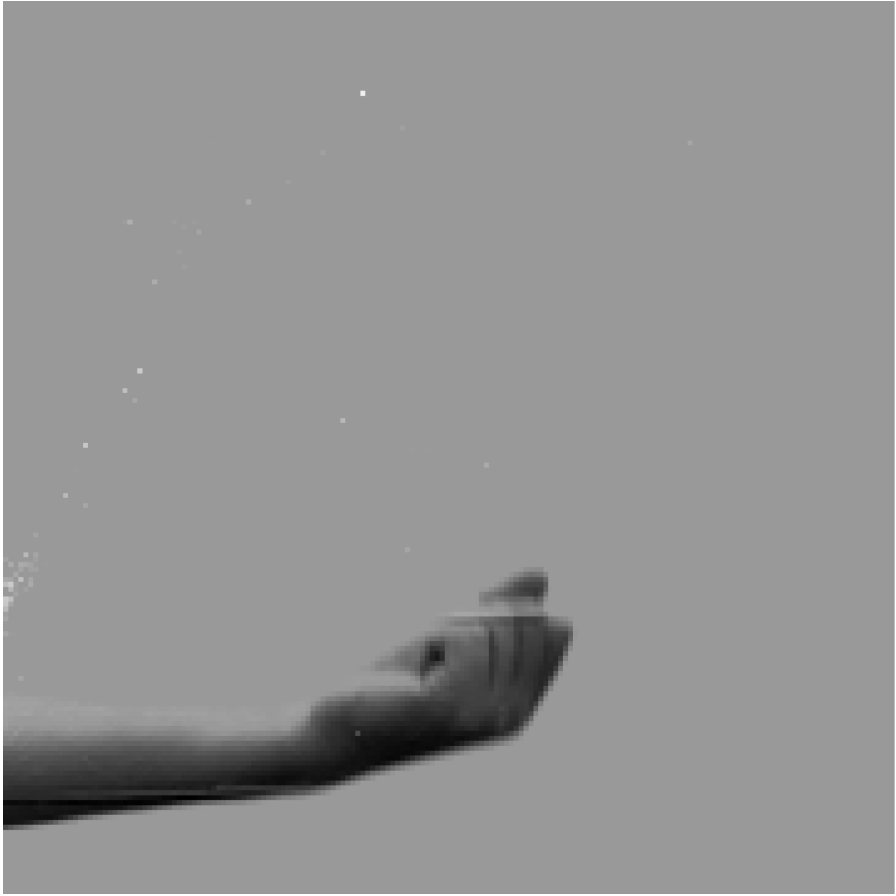}&
\includegraphics[width=0.19\textwidth]{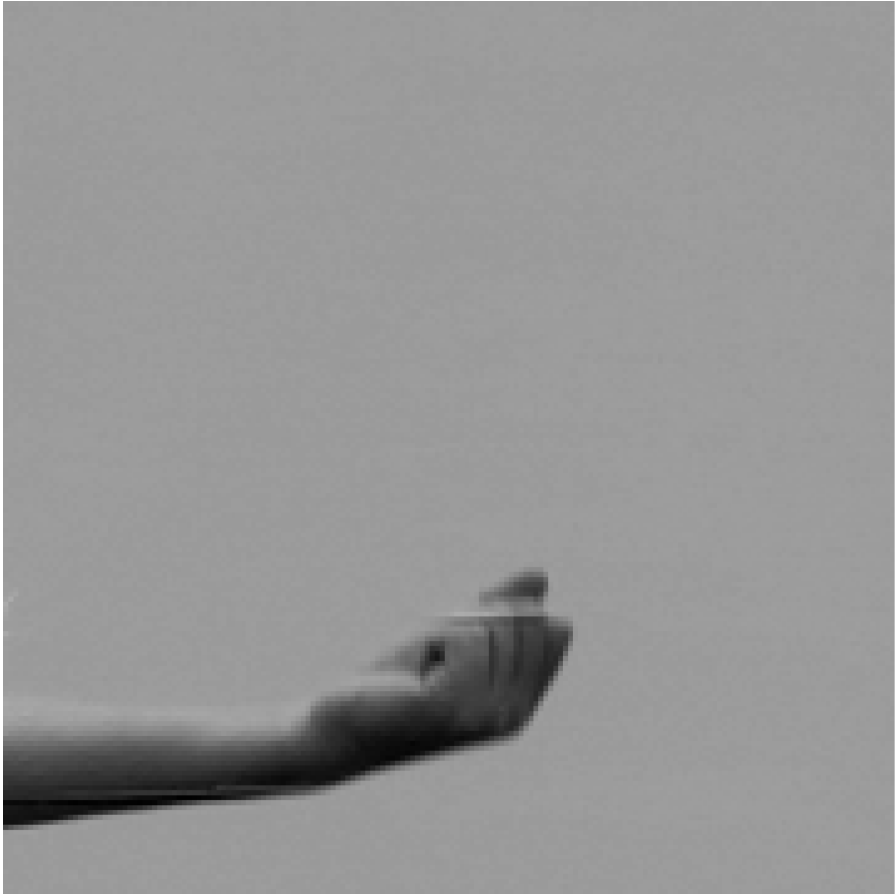}&
\includegraphics[width=0.19\textwidth]{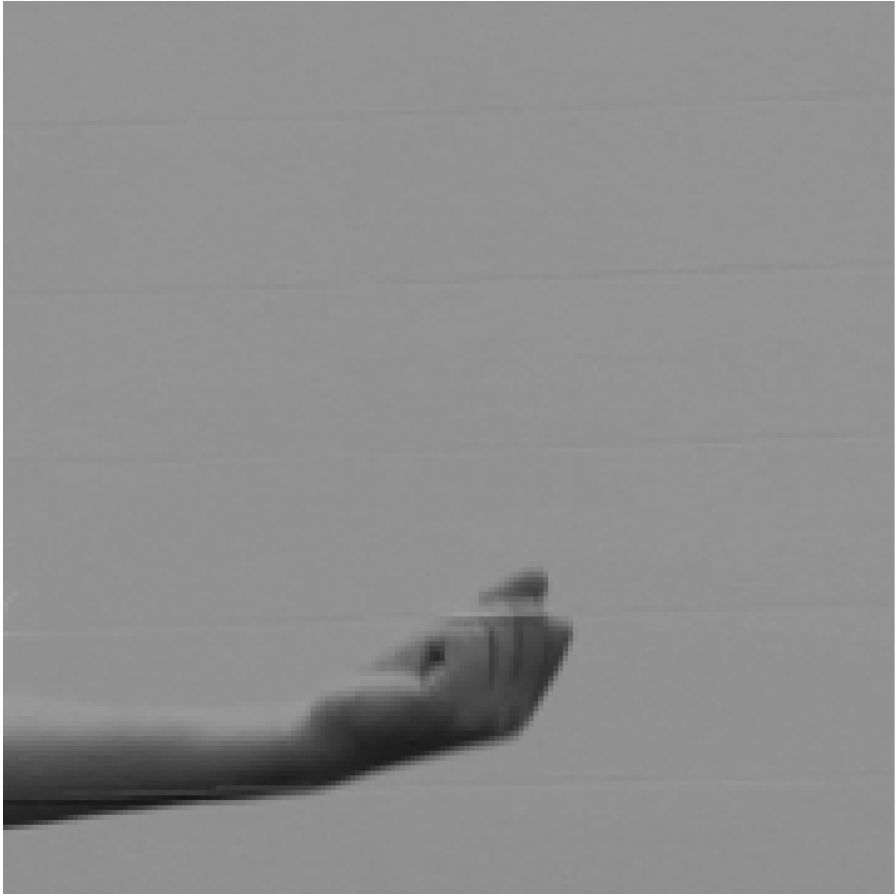}\\
\includegraphics[width=0.19\textwidth]{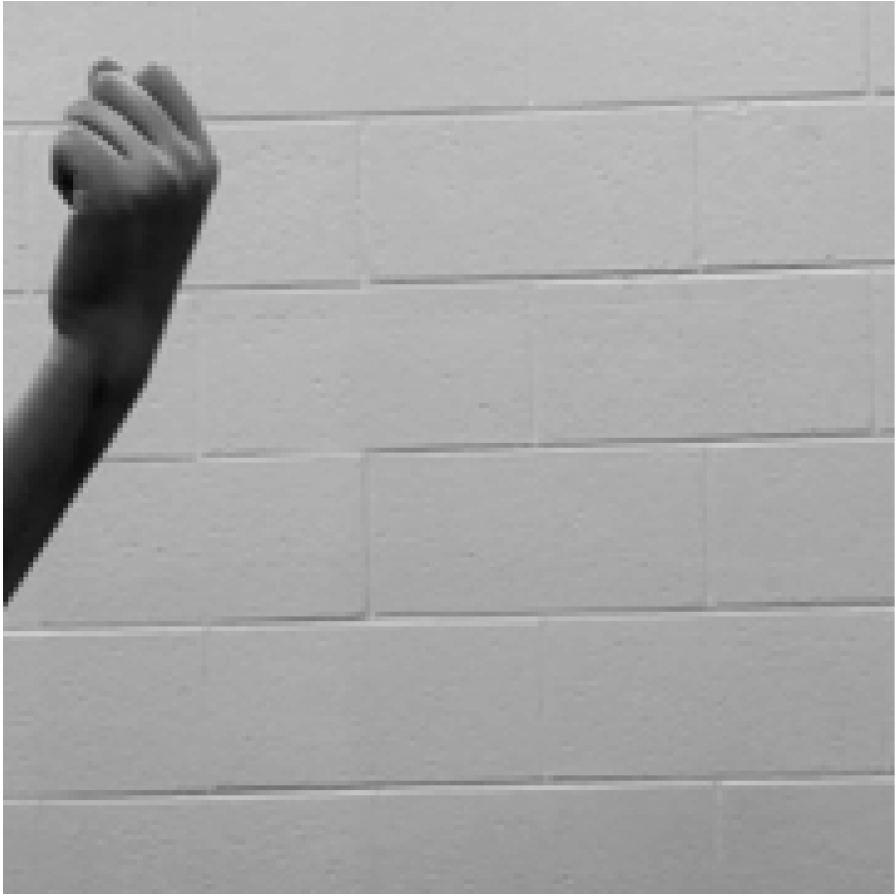}&
\includegraphics[width=0.19\textwidth]{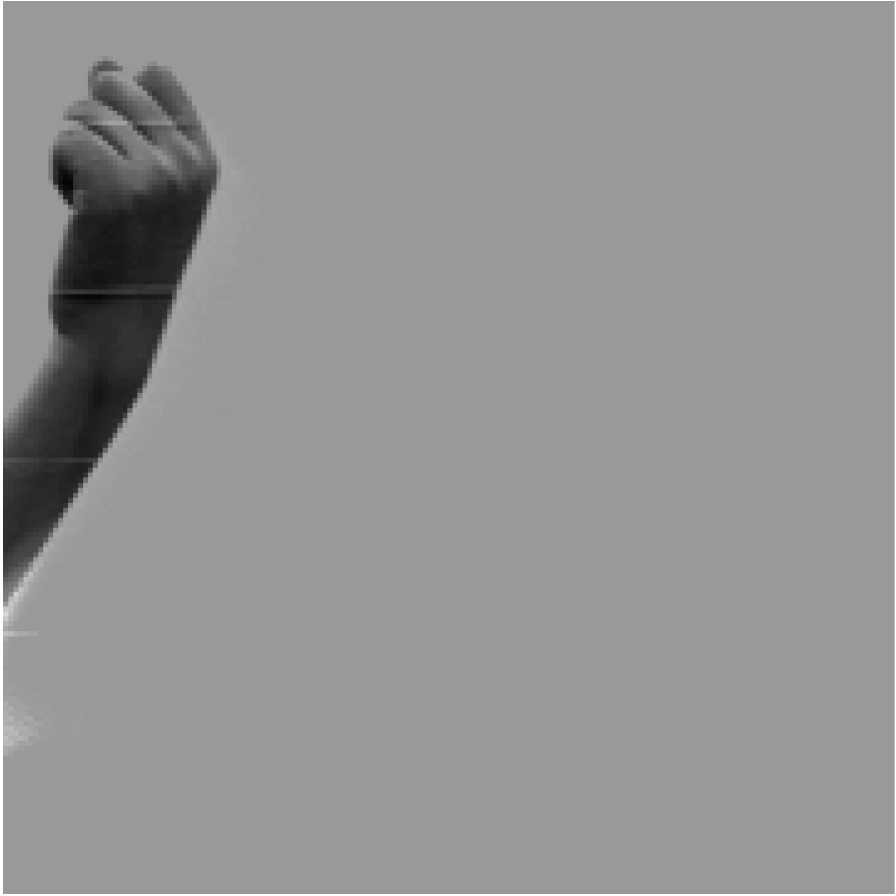}&
\includegraphics[width=0.19\textwidth]{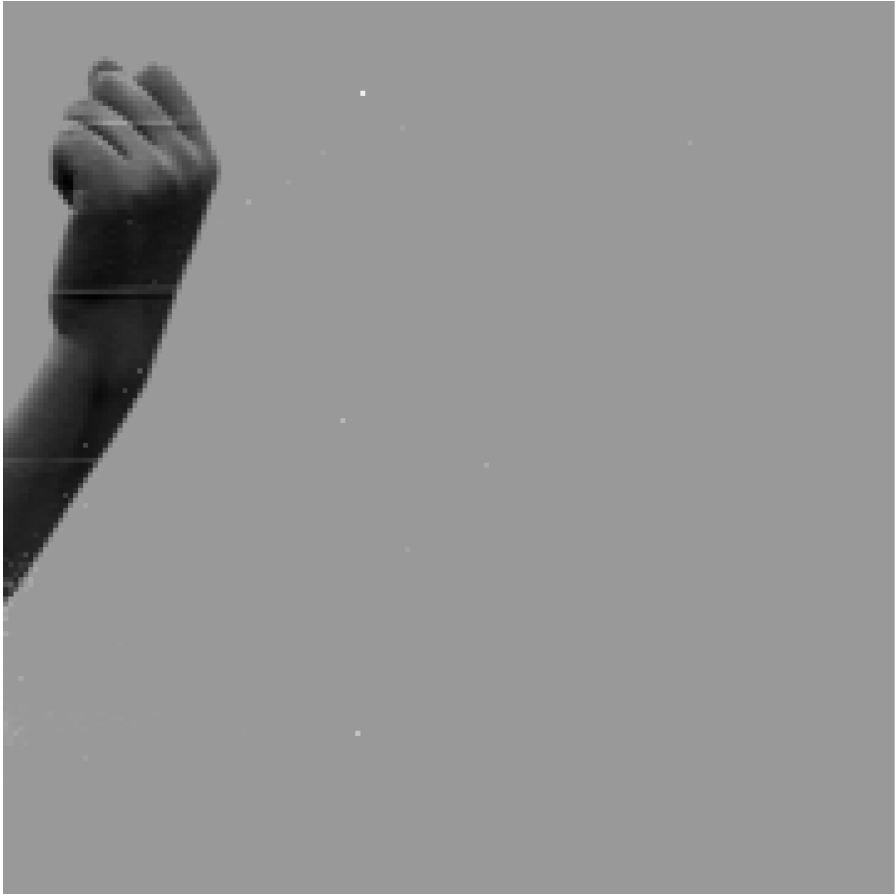}&
\includegraphics[width=0.19\textwidth]{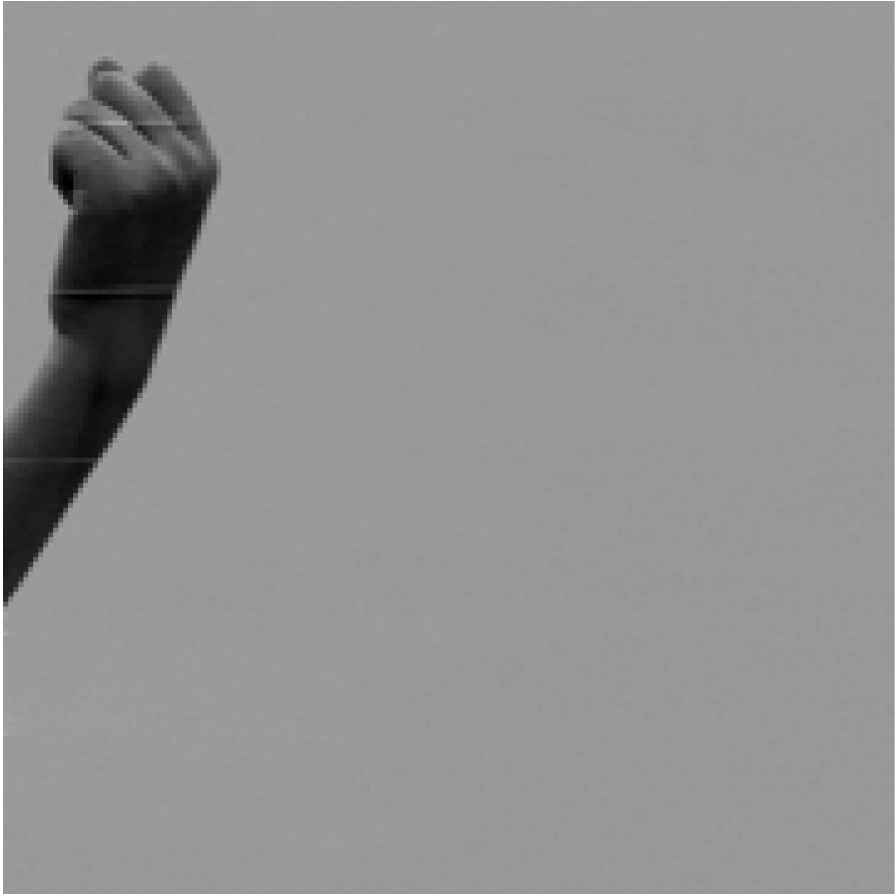}&
\includegraphics[width=0.19\textwidth]{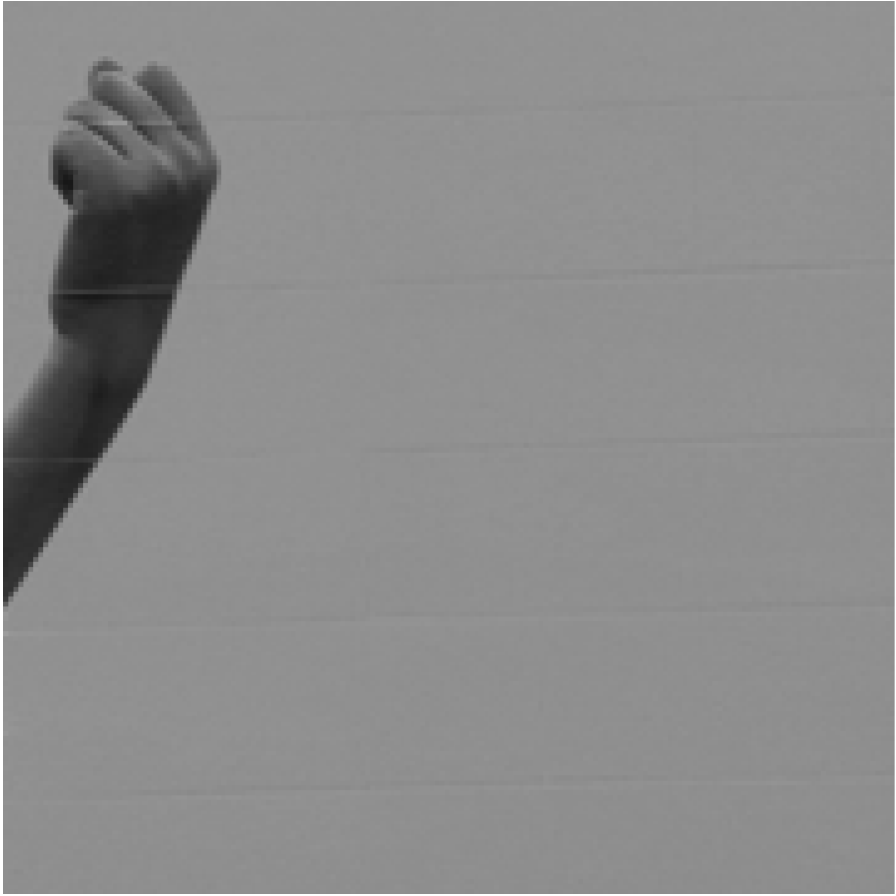}\\[-2pt]
Original frame &  LAGO &  SPCP & SPGL1 &  Alg.1
\end{tabular}
\vspace{-2pt}
\caption{Foreground detection for the arm motion video. The two rows correspond to the first and the last video frames, respectively.}\label{fig:exp2fg}
\end{figure*}

\begin{table}[h]
\centering
\caption{Quantitative comparison for the arm motion video}\label{tab2}
\vspace{-4pt}
\begin{tabular}{c|cc|ccc}
\hline \hline
    & RE & PSNR & Pr & Re & Fm\\ \hline
LAGO &0.0151&20.36&0.9617&0.8305&0.8913\\
SPCP& 0.0132&20.44&0.9666&0.8471&0.9029 \\
SPGL1 & 0.0132 & 35.65 & 0.9665 & 0.8610&0.9107 \\
Alg.~1 & 0.0104 & 35.67 & 0.9704 & 0.8234&0.8909\\
\hline\hline
\end{tabular}
\end{table}

\section{Conclusions and Future Work}\label{sec:con}
Moving object detection is one of the most fundamental tasks in video processing with a wide spectrum of applications, particularly in human-robot interaction. In the case of limited lightening conditions and/or time-varying illuminations, it becomes extremely challenging to separate a moving foreground with shadow from a static background. One classical type of methods is to segment each single frame into foreground and background. However, it usually loses the temporal smoothness and suffers from the intensive computation. In this work, we propose a novel dual-graph regularized motion detection approach. Specifically, we exploit the spatiotemporal geometry of the foreground by constructing the spatial and the temporal graph Laplacians, and adopt a weighted nuclear norm based regularization to utilize adaptive low-rankness of the background. The proposed algorithm is derived by applying the ADMM framework. Numerical results have shown our proposed method performs well on realistic data sets. In the future, we will develop fast methods based on the low-rank tensor decompositions and separate the shadow from the detected moving object under more sophisticated lightening environments. 



%

\section*{ACKNOWLEDGMENTS}

The research of Qin is supported by the NSF grant DMS-1941197.


\bibliographystyle{IEEEtran}
\bibliography{ref}

\end{document}